\PassOptionsToPackage{table,dvipsnames}{xcolor}
\documentclass[10pt]{visaresearch}

\usepackage{tabularx}
\usepackage{amsmath}
\usepackage{amssymb}
\usepackage{wrapfig}
\usepackage{makecell}
\usepackage{graphicx}

\newcommand\blfootnote[1]{%
  \begingroup
  \renewcommand\thefootnote{}%
  \footnote{#1}%
  \endgroup
}

\title{VRLLM: Turning LLMs into Powerful Tabular Classifiers}

\author{Yingtong Dou\textsuperscript{*}, Shubham Jain\textsuperscript{*}, Junpeng Wang, Hadi Abdullah, \\Vineeth Rakesh Mohan, Yuzhong Chen, Liang Wang, Yiwei Cai}

\affiliation{\textbf{Visa Research}}
\contribution{\textsuperscript{*}Equal contribution.}

\date{\today}
\correspondence{Yiwei Cai \email{yicai@visa.com}}

\abstract{
Large Language Models (LLMs)  have shown strong performance on text, but whether pretrained LLMs are effective for large-scale tabular classification remains unclear, as many previous works demonstrate that LLMs' tabular data classification performance falls behind tree-based models or neural architectures designed for tabular data.
At Visa Research, we managed to fine-tune a pretrained Llama 3.2 1B LLM on billions of rows of labeled tabular data and make it outperform a wide array of tabular classifiers and even our competitive production deep learning model.
For ease of reference, we refer to the model as the Visa Research LLM (\textbf{VRLLM}).
In this paper, we share how we achieved this exciting result, covering data preparation, training recipe, ablation studies of design choices and deployment validation.
We also benchmark frontier LLMs on the same task and analyze their prediction behavior across model families and generations.
Our interpretability experiments suggest that VRLLM learns feature-specific predictive associations, while component ablations indicate that the earliest two FFN layers contribute to the emergence of linearly decodable task information.
Our paper provides an engineering recipe and a focused empirical case study on adapting compact LLMs to industrial tabular prediction tasks.

}

\begin{document}

\maketitle

\begin{figure}[h]
    \centering
    \includegraphics[width=1.0\linewidth]{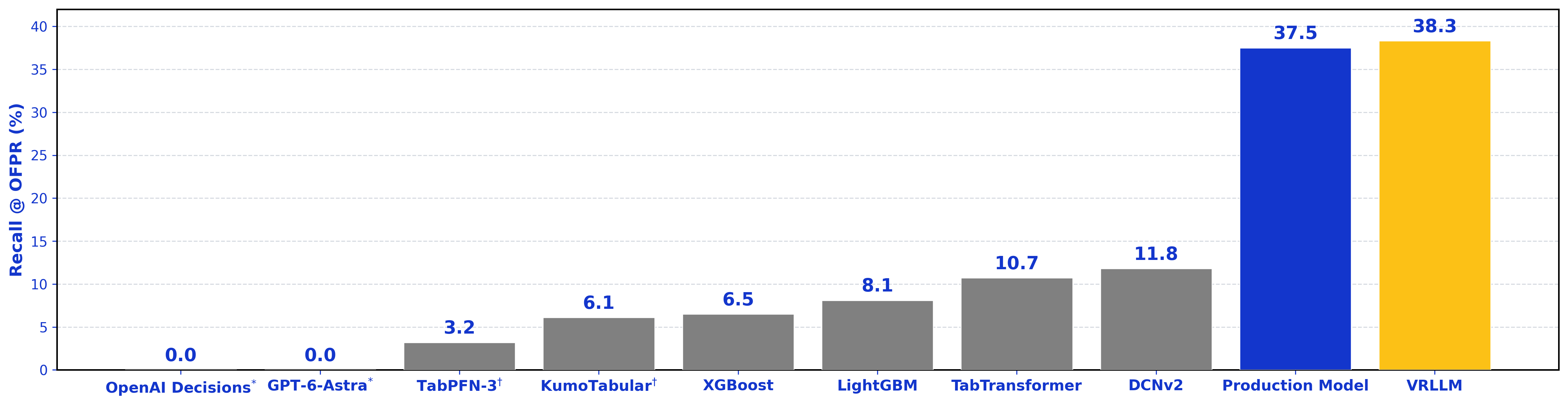}
    \caption{Classification performance of frontier models, tabular classifiers, the Visa production model, and VRLLM. Performance is reported as recall at the selected operating false-positive rate (OFPR). All experiments were conducted on offline data in a non-production environment for research purposes. Models marked $\ast$ were evaluated zero-shot on 10{,}000 examples; all others were evaluated on a larger dataset. Models marked $\dagger$ were evaluated few-shot without fine-tuning, while the remaining models were trained or fine-tuned on the same source data. See Sec.~\ref{sec:experiments} for more evaluation details and performance comparisons.}
    \label{fig:benchmark}
\end{figure}

\blfootnote{\textbf{Disclaimer.}
All trademarks are the property of their respective owners, are used for identification purposes only, and do not necessarily imply product endorsement or
affiliation with Visa. Performance results are limited to the specific research task, evaluation dataset, model versions, and experimental settings described in this paper. They should not be interpreted as statements about the general capabilities, commercial suitability, or relative market standing of the referenced models, products, or organizations.
To comply with legal and privacy policies, we do not release the dataset or any model trained on it.}
\blfootnote{Preprint. Work in progress.}

\section{Introduction}
\label{sec:introduction}

Visa processed 257.5 billion transactions in fiscal year 2025~\cite{visa2025factsheet}, producing large volumes of labeled, heterogeneous tabular data.
Such data support predictive models for future transactions.
For predictive modeling and tabular data classification, feature engineering combined with tree-based ensembles remains the dominant solution, as such models are efficient to train, update, and inspect~\citep{grinsztajn2022tree,mcelfresh2023neural}.

With the growing popularity of the Transformer~\citep{vaswani2017attention} and the LLMs and the foundation models built upon it, Transformer-based models have increasingly been applied to tabular data~\citep{huang2020tabtransformer, hegselmann2023tabllm, qu2026tabiclv2}.
Prior work spans specialized architectures, foundation models, and LLM adaptation. TabTransformer and SAINT model heterogeneous feature interactions through attention~\citep{huang2020tabtransformer,somepalli2021saint}, while TabPFN~\citep{hollmann2025accurate} and TabICL~\citep{jingangtabicl} transfer predictive capabilities across tasks through in-context learning. Complementary approaches serialize tables for LLM-based classification~\citep{hegselmann2023tabllm,ruan2024language} or encode transaction histories using recurrent networks, Transformers, and generative pretraining~\citep{skalski2023towards,zhang2023fata, dou2025transactiongpt}. More recently, LLM4ES~\citep{shestov2025llm4es} and SeqLLM~\citep{li2026seqllm} adapt pretrained LLMs to behavioral sequences through textual event descriptions and specialized event tokenization, respectively. 

Building on these directions, we investigate \textbf{whether a compact pretrained causal LLM, fine-tuned directly on serialized transaction records, can serve as the classifier itself using its existing tokenizer and prediction head, without a specialized transaction encoder or tokenizer}.
The rationale for this approach is threefold:

\begin{itemize}
    \item Our data contain many categorical fields, some with extremely high cardinality and rich semantic information (e.g., more than 50M unique merchants). Assigning a separate embedding to each category can incur substantial memory and computational costs and provides limited support for unseen categories. In contrast, a pretrained LLM tokenizer uses a fixed vocabulary to represent category values compositionally, enabling it to encode unseen values and potentially capture semantic similarities among categories.

    \item Large-scale labeled transaction data provide domain-specific supervision beyond the general knowledge acquired during pretraining. Fine-tuning can encode task-relevant transaction patterns and feature--label relationships into model parameters, enabling predictions on new records without requiring labeled examples at inference.

    \item Our Transformer-based payment foundation models~\citep{zhang2023fata,dou2025transactiongpt,yeh2026treasure} perform well on transaction classification, motivating us to investigate whether pretrained LLMs with similar backbones can achieve comparable results.
\end{itemize} 
We therefore prioritize domain-supervised fine-tuning to combine pretrained semantic representations with predictive knowledge learned from labeled payment data. 

Our results demonstrate that a pretrained Llama 3.2 1B model~\citep{dubey2024llama}, fine-tuned on billions of serialized transaction records, can achieve strong classification performance (Fig.~\ref{fig:benchmark}).
VRLLM achieves 38.3\% recall at the operating false-positive rate (OFPR) and 35.3\% PR-AUC, exceeding the production-model baseline by 0.8 and 0.1 percentage points, respectively.
In addition to comprehensive benchmarking and ablation studies, we conduct interpretability analyses to examine how pretraining and domain-specific learning contribute to the LLM's predictions (Sec.~\ref{sec:interpretation}). We make four main contributions:

\begin{itemize}
    \item \textbf{Effective domain adaptation at scale.}
    We show that a general pretrained LLM can outperform specialized tabular classifiers on our proprietary benchmark. We provide an end-to-end training recipe combining serialization, next-token supervision, parameter-efficient adaptation, and optimization for imbalanced data.

    \item \textbf{Analysis of performance and robustness.}
    We examine how input semantics, pretrained knowledge, and architectural choices affect VRLLM's classification performance and evaluate VRLLM's robustness under deployment-oriented conditions alongside the production model.

    \item \textbf{Feature-specific priors and early-layer contributions.}
    Our interpretability analyses suggest that VRLLM combines multiple sources of predictive knowledge: product codes provide priors grounded in training data, while dollar buckets provide ordinal information. Layer-wise probing further suggests that early feed-forward networks play an important role in encoding domain-specific knowledge.

    \item \textbf{Evaluation of frontier LLMs and decision models.}
    We evaluate frontier LLMs and decision models without domain-specific adaptation. Although newer model generations generally achieve better overall classification performance on our benchmark, their recall at OFPR remains at 0.00\%, falling well short of production performance.
\end{itemize}

\section{Task and Data}

We first define the binary transaction-classification task, summarize the dataset and temporal split, and describe the tabular data serialization method.
To comply with company policies, we omit the raw feature values, exact feature definitions, calendar dates, and operating FPR; we report all other details needed to interpret the experiments.

\subsection{Task Definition}

We study binary classification.
Given historical labeled tabular data, our objective is to train a binary classifier that predicts the probability that a tabular record in the test data belongs to the positive class.
The validation and test data occur later in time than the training data. 
The data is highly imbalanced, with an extremely rare positive class.
Under company policies, we cannot disclose the exact size of the data, positive class definition, class rate, and training/validation/test time splits.

\subsection{Data Overview}

The tabular data include hundreds of categorical and numerical features derived from payment transactions.
The feature set used in this study excludes personally identifiable information (PII) under our internal data-governance definition.
These features have been used to train non-LLM-based tabular classifiers and have proven effective.
To maximize the utility of the tabular data for LLMs, we use the original textual values instead of categorical codes for features with semantic meaning.
For instance, we use merchant-name string instead of ID to represent merchants.

In contrast to many public tabular data benchmarks and industry-scale tabular datasets~\citep{mcelfresh2023neural}, our dataset is significantly larger, containing billions of rows with high-fidelity labels.
Another property of the data is the presence of high-cardinality fields such as the merchant-name field, which has more than 50M unique values.

\begin{figure}
    \centering
    \includegraphics[width=0.98\linewidth]{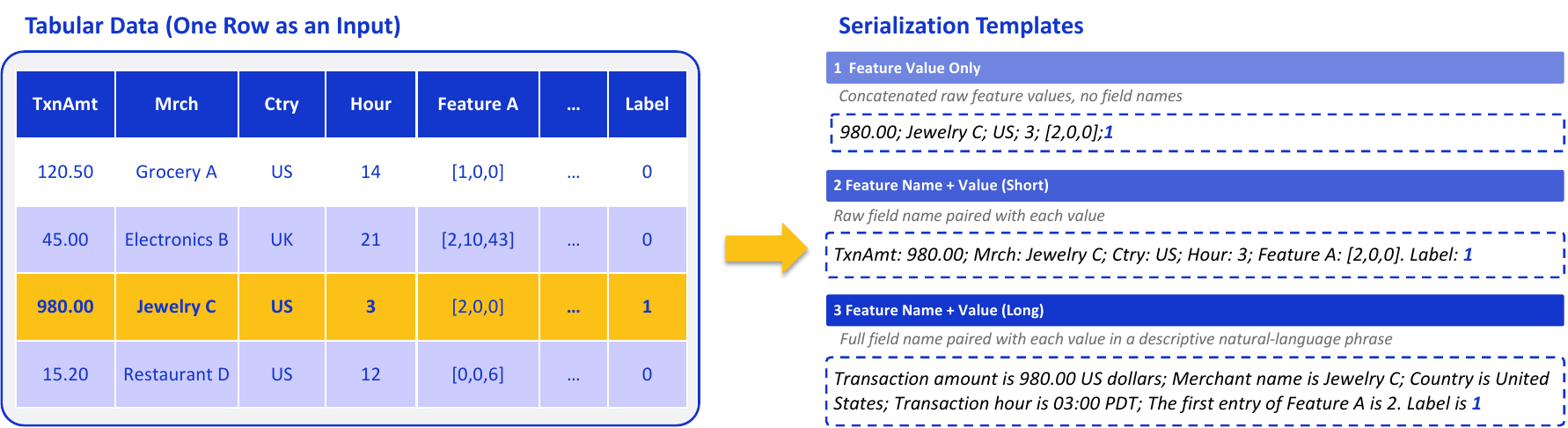}
    \caption{Three different tabular data serialization templates. The data and templates are used for illustration purposes and do not reflect the appearance or contents of the production data. The label token is masked during fine-tuning and inference.}
    \label{fig:text_template}
\end{figure}

\subsection{Serialization Method}

Building on prior serialization work~\citep{ruan2024language}, we construct three fixed-order text representations that vary in whether and how they include field names (Fig.~\ref{fig:text_template}). We follow three design principles:

\textbf{(1) Enriching the semantics of serialized tabular data.}
For features and fields whose raw values include text, we retain the text instead of processed codes or values used by non-LLM ML models.
For features with complex aggregation rules, we expand their names to make them easier to understand.
We also design text templates tailored to the properties of each feature to maximize their semantic richness. For example, for highly sparse feature vectors with zero-value entries, we directly state that a feature is zero or null instead of describing each entry of the vector.

\textbf{(2) Reducing template length by removing redundant information.} Since fine-tuning on a large serialized dataset is costly, we remove unnecessary tokens from templates to shorten sequences and lower training costs.
We use concise text templates to describe each feature name and its value.
For numerical features, we convert floating-point numbers to integers if they are derived from integers; otherwise, we round the values to six decimal places and scale them to integers.

\textbf{(3) Designing different templates based on semantic richness.} 
Since it is unclear how much the semantic content of field names and textual values contributes to the predictive modeling task, we design three different versions of serialization templates based on their semantic richness (Fig.~\ref{fig:text_template}).
\textit{Feature Value Only} is the template in which only raw feature values are presented;
\textit{Feature Name + Value (Short)} is the template that includes feature names and feature values with minimal transformation;
\textit{Feature Name + Value (Long)} is the most complex template, transforming each feature based on its meaning and data properties.

All templates preserve the same fixed field order and contain identical underlying feature values.
After applying the above principles, each row of tabular data is transformed into text data ranging from 600 to 1,300 tokens.
During tokenization, we use left padding and truncation with a maximum input length of 4,096 tokens.

\textbf{Note on numerical value tokenization.}
Under the serialization method described in (2), numerical features are converted to integers and tokenized as sequences of digit strings.
Unlike numerical feature embedding methods that explicitly encode numerical structure~\citep{gorishniy2022embeddings}, this tokenization scheme does not explicitly preserve the ordering of the original values.
For example, after conversion to the integer string \texttt{3125}, the floating-point value \texttt{3.125} is tokenized into \texttt{312} and \texttt{5} by the Llama 3 tokenizer~\citep{dubey2024llama}.
Nevertheless, numerical features in our datasets have relatively few distinct values compared with high-cardinality categorical features.
We hypothesize that this lower cardinality, together with the fixed-order template and large scale of our training data, enables the LLM to learn relationships among numerical values.
The ordinal-based prior observed across dollar-buckets in Sec.~\ref{sec:data_interpretation} further suggests that the LLM captures some ordinal relationships despite representing numerical values as sequences of digit tokens.

\section{Training Recipe}
\label{sec:training}

\begin{table}[t]
\centering
\caption{Training configuration for VRLLM.}
\label{tab:training-configuration}
\footnotesize
\renewcommand{\arraystretch}{1.08}
\begin{tabularx}{0.7\linewidth}{
    @{}
    >{\raggedright\arraybackslash}p{0.25\linewidth}
    >{\raggedright\arraybackslash}X
    @{}
}
\toprule
\textbf{Setting} & \textbf{Value} \\
\midrule
Backbone
    & Llama 3.2 1B (LoRA: rank=32, alpha=64, dropout=0.1) \\
\# Trainable parameters
    & 22.5M (1.8\%) \\
Effective batch size
    & 32 (gradient accumulation - 512) \\
Optimizer / LR / weight\_decay
    & AdamW / $2.0 \times 10^{-5}$ / $1.0 \times 10^{-2}$ \\
LR scheduler
    & cosine, linear warmup = 10k steps \\
Base model precision / parallelism
    & BF16 / Distributed Data Parallel (DDP) \\
Hardware config 
    & 4 × 80-GB NVIDIA A100/H100 GPUs \\
\bottomrule
\end{tabularx}
\end{table}

To fine-tune a pretrained LLM for text classification, there are two common approaches: fine-tuning an encoder-only or decoder-only LLM.
We found that pretrained encoder-only Transformers performed substantially worse than decoder-only LLMs under our preliminary 1M-example configuration (see Table~\ref{tab:embedding-model-ablation}); therefore, we focus on fine-tuning decoder-only LLMs.
Our detailed training recipe for decoder-only LLMs is described below.

\textbf{Classification as next-token prediction.}
We follow the autoregressive formulation of decoder-only LLMs and fine-tune the model to predict a label token after each serialized record.
Given a serialized sequence $\mathbf{x} = (x_1, x_2, \ldots, x_{T-1}, x_T)$ where $x_T$ is the ground-truth label token (0 or 1), the training loss is the cross-entropy computed only for the last label token, with all other tokens excluded from the loss:

\begin{equation}
\mathcal{L} = - \log \frac{\exp(z_{T-1, x_T})}{\displaystyle\sum_{v=1}^{V} \exp(z_{T-1, v})},
\end{equation}

where $z_{T-1} = f_\theta(x_1, x_2, \ldots, x_{T-1}) \in \mathbb{R}^{V}$ is the logit vector produced by the causal language model $f_\theta$ for predicting the next token $x_T$ and $V$ is the vocabulary size of the corresponding LLM being fine-tuned.

The loss function above essentially optimizes the probability that the next token is the ground-truth class (0 or 1) given the serialized tabular data as input.
This objective reuses the pretrained tokenizer, token embeddings, decoder, and LM head.
Apart from LoRA modules inserted into selected decoder projections, it adds no field-specific input encoder, vocabulary extension, or task-specific classification head.
We use this design to test whether the existing language-model interface can support serialized tabular classification.

At inference time, given an input sequence without the label token $\mathbf{x}' = (x_1, x_2, \ldots, x_{T-1})$, the renormalized positive-class probability is computed as follows:

\begin{equation}
s(\mathbf{x}') = \frac{\exp(z_{T-1, \texttt{1}})}{\exp(z_{T-1, \texttt{0}}) + \exp(z_{T-1, \texttt{1}})}.
\end{equation}
Because evaluation requires a continuous score rather than generated text, we compute s(x') directly from the LM-head logits for the tokens representing labels 0 and 1. We then calibrate this score using our business-specific method.

\textbf{LoRA configuration.}
Since our serialized tabular inputs are relatively long (600--1.3k tokens each) and our training infrastructure has limited capacity (as shown in Table~\ref{tab:training-configuration}), we adopt low-rank adaptation (LoRA)~\citep{hu2022lora} to improve training efficiency.
Specifically, the embedding layer and LM head (tied in Llama 3.2 1B) are frozen during fine-tuning and LoRA adapters are only applied to the attention layers (\texttt{q\_proj}, \texttt{k\_proj}, \texttt{v\_proj}, \texttt{o\_proj}) and FFN projections (\texttt{gate\_proj}, \texttt{up\_proj}, \texttt{down\_proj}).
A more detailed configuration is presented in Table~\ref{tab:training-configuration}.
Our subsequent  interpretability analysis of these layers is presented in Sec.~\ref{sec:model_interpretation}.

\textbf{Data sampling.} The long input sequences  limit the per-GPU microbatch size to at most 8 on a GPU with 80 GB of VRAM.
Given the extreme class imbalance in the unsampled training data, such small batches may contain few or no positive examples, resulting in noisy, negative-class-dominated gradients.
To address this challenge, we retain all positive-class records while downsampling the negative-class records to obtain positive-to-negative ratios ranging from 1:3 to 1:10.

\textbf{Gradient accumulation.}
To retain sufficient negative examples while maintaining training stability,
we use a staged fine-tuning strategy: the LLM begins with a batch size of 32, and gradient accumulation is introduced after fine-tuning on approximately 100 million unique training records.
An accumulation factor of 16 (corresponding to an effective batch size of 512) achieves the best performance by improving the consistency of minority-class representation per update, gradient stability and optimizer-update frequency~\citep{imbalanced_data_2009, mccandlish2018empiricalmodellargebatchtraining}.

We made the following observations while tuning batch size:
small batches enable frequent updates early in training, but continuing with a batch size of 32 eventually leads to slower improvement and lower performance.
A gradient accumulation factor of 8 performs well but is slightly inferior to 16, while a factor of 80 performs worse.
The poorer performance with a factor of 80 may partly reflect its substantially lower gradient update frequency.
Therefore, an intermediate effective batch size stabilizes the minority-class signal without excessively reducing update frequency.

\textbf{Training optimizations.} We replaced the fixed CUDA memory-split configuration with expandable segments and disabled CUDA debugging settings to reduce memory fragmentation and synchronization overhead. We also increased the number of DataLoader workers from 0 to 4 and enabled persistent workers to reduce data-loading latency and GPU idle time. Collectively, these optimizations reduced peak GPU memory usage by approximately 50\% and training time by approximately 30\%, enabling us to maintain the same batch size with half as many GPUs.

\section{Experiments}
\label{sec:experiments}

First, we compare the performance of VRLLM with that of different types of tabular classifier baselines on a large-scale temporal holdout test set (Sec.~\ref{sec:overall_results}). Then, we benchmark the zero-shot classification performance of frontier LLMs on a 10,000-transaction subset (Sec.~\ref{sec:zeroshot_exp}), conduct controlled ablations (Sec.~\ref{sec:ablation}) and assess temporal stability, unseen-merchant performance, and ensemble complementarity (Sec.~\ref{sec:deploy}).

\subsection{Large-scale Evaluation}
\label{sec:overall_results}

\textbf{Experiment setup.}
We first compare the performance of VRLLM with representative tabular classification baselines on a large-scale temporal holdout test set.
\textbf{Recall@OFPR} is a business metric that measures the recall for the positive class at a proprietary operating false-positive rate.
The area under the precision-recall curve (\textbf{PR-AUC}) is a metric for measuring performance in  imbalanced binary classification.
All models' prediction scores are calibrated using the same internal method before computing the business metric.
Note that the temporal holdout test set was curated specifically for this study; therefore, while the absolute metric values do not reflect real-world performance, they support valid relative comparisons.
Fig.~\ref{fig:benchmark} reports the same Recall@OFPR results as Table~\ref{tab:main_results}.

Our baselines include classic tree-based models (XGBoost~\citep{chen2016xgboost}, LightGBM~\citep{ke2017lightgbm}), tabular neural networks (DCNv2~\citep{wang2021dcnv2}, SAINT~\citep{somepalli2021saint}, TabTransformer~\citep{huang2020tabtransformer}), emerging tabular foundation models (TabPFN-3~\citep{grinsztajn2026tabpfn3}, TabFM~\citep{kong2026tabfm}, TabICLv2~\citep{qu2026tabiclv2}, and KumoTabular~\citep{nvidia2026kumotabular}), and our proprietary production model, which was designed and trained on the same labeled data under the same time split for the same task.
Model sources and license information can be found in Table~\ref{tab:model-sources}.

Among these models, only the tabular foundation models are neither trained nor fine-tuned on our data.
Instead, we leverage their in-context learning (ICL) capabilities to perform inference on the test data using the same set of 800K transactions as context (the support set).
This is the largest common support-set size that fits on a single GPU with 80 GB of VRAM for all evaluated tabular foundation models. These transactions are randomly sampled from the training data at a positive-to-negative class ratio of 1:1.
The other baselines are trained on the same set of transactions as VRLLM.
For each trainable model, we carry out a hyperparameter search with a fixed budget and report the best performing configuration.

To make a fair comparison, we align the number of parameters of three tabular neural networks to VRLLM's 1B parameters. For these neural baselines, we use EmbeddingBag~\citep{pytorch_embeddingbag} to handle high-cardinality categorical features. At test time, unseen categories are represented by the mean embedding in the corresponding feature’s embedding table.
Both XGBoost and LightGBM are trained directly on the integer-coded feature matrix, in which every categorical field is already mapped to a non-negative id.
For XGBoost we use \texttt{tree\_method="hist"} and treat these columns as ordinal numeric features, so splits take the form $x \leq \theta$.
For LightGBM we instead pass the categorical column indices via the \texttt{categorical\_feature} argument, enabling its native Fisher-style optimal partitioning that produces true set-valued splits. No additional encoding (one-hot, target, or hashing) is applied to either model.
All models are evaluated on the same holdout test set.

\begin{table}[t]
    \centering
    \caption{Performance on a large-scale temporal holdout test set.
    Models are grouped by whether they undergo task-specific training
    or fine-tuning. Bold and underlined values indicate the best and
    second-best results, respectively.}
    \label{tab:main_results}
    \resizebox{0.98\linewidth}{!}{%
        \setlength{\tabcolsep}{5pt}
        \renewcommand{\arraystretch}{1.15}
        \begin{tabular}{l*{12}{c}}
            \toprule
            & \multicolumn{4}{c}{\textbf{No task-specific training}}
            & \multicolumn{8}{c}{\textbf{Task-trained or fine-tuned}} \\
            \cmidrule(lr){2-5} \cmidrule(lr){6-13}
            \textbf{Model}
            & TabPFN-3
            & TabFM
            & TabICLv2
            & KumoTabular
            & XGBoost
            & LightGBM
            & SAINT
            & TabTransformer
            & DCNv2
            & \makecell[c]{Production \\ model}
            & VRLLM
            & \makecell[c]{Production \\ + VRLLM} \\
            \textbf{Reference}
            & \citep{grinsztajn2026tabpfn3}
            & \citep{kong2026tabfm}
            & \citep{qu2026tabiclv2}
            & \citep{nvidia2026kumotabular}
            & \citep{chen2016xgboost}
            & \citep{ke2017lightgbm}
            & \citep{somepalli2021saint}
            & \citep{huang2020tabtransformer}
            & \citep{wang2021dcnv2}
            & --- & --- & --- \\
            \textbf{\# Parameters}
            & 53M & 410M & 27M & 213M & 500K & 130K
            & 1B & 1B & 1B & --- & 1B & --- \\
            \midrule
            Recall@OFPR (\%) $\uparrow$
            & 3.2 & 5.3 & 4.0 & 6.1 & 6.5 & 8.1
            & 7.3 & 10.7 & 11.8 & 37.5
            & \underline{38.3} & \textbf{45.9} \\
            PR-AUC (\%) $\uparrow$
            & 8.2 & 6.5 & 8.4 & 11.2 & 12.7 & 13.1
            & 14.0 & 15.7 & 18.8 & 35.2
            & \underline{35.3} & \textbf{38.7} \\
            \bottomrule
        \end{tabular}%
    }
\end{table}

\textbf{Performance comparison.}
From Table~\ref{tab:main_results}, we make the following observations:
\begin{itemize}
    \item Fine-tuned Llama 3.2 1B (VRLLM) achieves the best performance among all individual models, outperforming the production model by 0.8 points in Recall@OFPR. This result demonstrates that an LLM-based approach retaining the original vocabulary, tokenizer, and language modeling head can be competitive for tabular predictive modeling.
    
    \item SAINT and TabTransformer are Transformer-based models designed for tabular data, yet both substantially underperform VRLLM. One possible explanation is that categorical embedding tables account for a large proportion of their parameters, while learning effective embeddings for infrequent categories in the long tail remains challenging. We hypothesize that the tokenization and representation of tabular fields play a critical role in performance on our task. The experiments in Sec.~\ref{sec:ablation} and Sec.~\ref{sec:interpretation} further investigate the advantages of fine-tuned LLMs over tabular Transformers in this setting.
    
    \item The tabular neural networks outperform the tree-based models, potentially reflecting their ability to capture interactions among the rich and informative categorical features in our datasets. Although the tree-based models support native categorical splits, this capability may be less effective for extremely high-cardinality features, particularly when many categories have few observations. These results suggest that categorical feature representation and interaction modeling may contribute to the observed performance gap.
    
    \item The pretrained tabular foundation models underperform all supervised models. This gap may reflect limitations in context capacity and the complexity of our classification task: a limited set of in-context examples may not adequately represent the variability in the training data. These results suggest that domain-specific fine-tuning may help these models learn predictive patterns that are difficult to infer from in-context examples alone.
\end{itemize}

The above evidence underscores the importance of  feature and model-architecture engineering for strong performance on this decision-critical application.
Directly adopting pretrained tabular foundation models and classic tabular classifiers is suboptimal.
Compared with the production model, which is purpose-built for the classification task, fine-tuning a pretrained LLM requires much less effort in feature engineering and model-architecture optimization.
This opens up new possibilities for applied machine learning in industry.
Meanwhile, after being fine-tuned on proprietary data, the LLM becomes a tabular classifier (or decision model) that offers desirable properties such as a bounded single-score output and a relatively small model size.

\subsection{Zero-shot Performance of Frontier Models}
\label{sec:zeroshot_exp}

\begin{wrapfigure}{l}{0.51\textwidth}
    \centering
    \includegraphics[width=\linewidth]{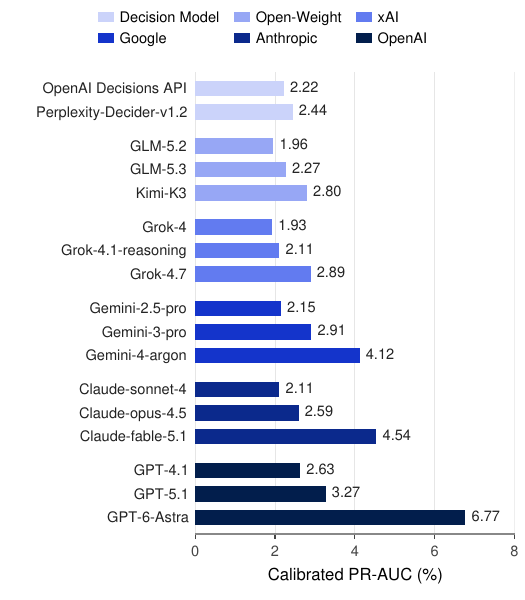}
    \caption{Zero-shot classification performance on a 10,000-transaction subset
    across frontier model families and generations.}
    \label{fig:frontier}
\end{wrapfigure}

Besides fine-tuning models or making predictions using ICL as described above, we also test the zero-shot classification performance of frontier LLMs using the Feature Name + Value (Long) template, as this template conveys more context about each input field.
Due to high API costs, we benchmark frontier LLMs on a subset of 10,000 transactions sampled from the holdout test set described above, with a 1:9 ratio of positive to negative examples.
This experiment evaluates the usefulness of frontier LLMs’ prior knowledge, reasoning, and zero-shot tabular classification capability for our proprietary classification task.
Except for the 28B open-weight Perplexity-Decider-v1.2~\citep{pplx_decider_v1_2_2026}, which was hosted on our on-prem server, all other models were accessed through secure enterprise LLM APIs.
For frontier LLMs, we instruct them to generate a score between 1 and 1000 for binary classification.
For Perplexity-Decider-v1.2, we extract the two class logits from its readout head and compute the positive-class probability using the default temperature.
The OpenAI Decisions API~\cite{openai_decisions_api_2026} directly returns the probability of the positive outcome.
We convert and calibrate these scores and probabilities before computing PR-AUC (comparable to the numbers reported in Table~\ref{tab:main_results}).

All frontier models achieve \textbf{zero} Recall@OFPR, indicating that their performance remains well below production requirements.
Nevertheless, Fig.~\ref{fig:frontier} shows consistent improvements in PR-AUC across successive generations of frontier models.
In particular, GPT-6-Astra substantially outperforms earlier generations of frontier LLMs.

\textbf{Behavioral analysis across frontier models.}
To understand the score discrepancy and prediction behavior of different models, we jointly analyze the inputs and outputs of these models and summarize the resulting observations below:                            
  \begin{itemize}                                
  \item \textbf{Top-ranked models show better base-rate calibration.} Top-3-ranked models output finely graded scores with means $0.06$--$0.11$, closely matching the positive-class base rate.
  Mid-ranked models (e.g., Gemini-3-pro, Grok-4) exhibit strongly bimodal distributions, with medians near $0.02$ but 90th percentiles above $0.85$. Bottom-ranked models collapse onto a narrow high-prior band: Gemini-2.5-pro has median $0.965$ and Perplexity-Decider-v1.2 confines $99.9\%$ of scores to $[0.59, 0.69]$.

  \item \textbf{Output resolution matters alongside calibration.} GPT-6-Astra outputs $266$ distinct score values among 10K inputs, compared with $114$ for Claude-fable-5.1 and only $40$--$49$ for the OpenAI Decisions API, Claude-sonnet-4, and Claude-opus-4.5. Such coarse output grids restrict ranking granularity and can limit achievable PR-AUC, regardless of the underlying reasoning quality.                                                              
  \item \textbf{GPT-6-Astra separates the two classes most clearly.} It assigns an average score of $0.26$ to positive transactions versus $0.09$ to negative ones, a gap of $0.17$ on the $[0,1]$ scale. The corresponding gaps are $0.15$ for Claude-fable-5.1 and $0.15$ for Gemini-4-argon, but fall to only $0.09$ for Gemini-2.5-pro ($0.92$ versus $0.83$) and $0.01$ for Perplexity-Decider-v1.2 ($0.61$ versus $0.60$), whose compressed score distributions limit ranking information.
  
   \item \textbf{Top-ranked models encode useful merchant-category and transaction-amount priors.} The top-3 ranked models exhibit a stronger understanding of merchant-category priors than the remaining frontier models, consistent with the findings of our prior work~\citep{wang2026understandingllmevaluator}. Also, these models assign higher positive-class probabilities to larger-dollar transactions, mirroring the dollar-bucket-conditioned priors we identified in VRLLM (Sec.~\ref{sec:data_interpretation}). 
  \end{itemize}

The additional analysis of frontier models can be found in Appendix~\ref{appdix:frontier_model}.
The above evidence suggests that, while frontier models are improving in zero-shot classification on our domain-specific data, task-specific supervised training remains necessary for producing fine-grained prediction scores that capture domain-specific predictive signals.

\subsection{Ablation Study}
\label{sec:ablation}
\begin{figure}
    \centering
   \begin{minipage}[t]{0.48\linewidth}
        \centering
        \includegraphics[width=\linewidth]{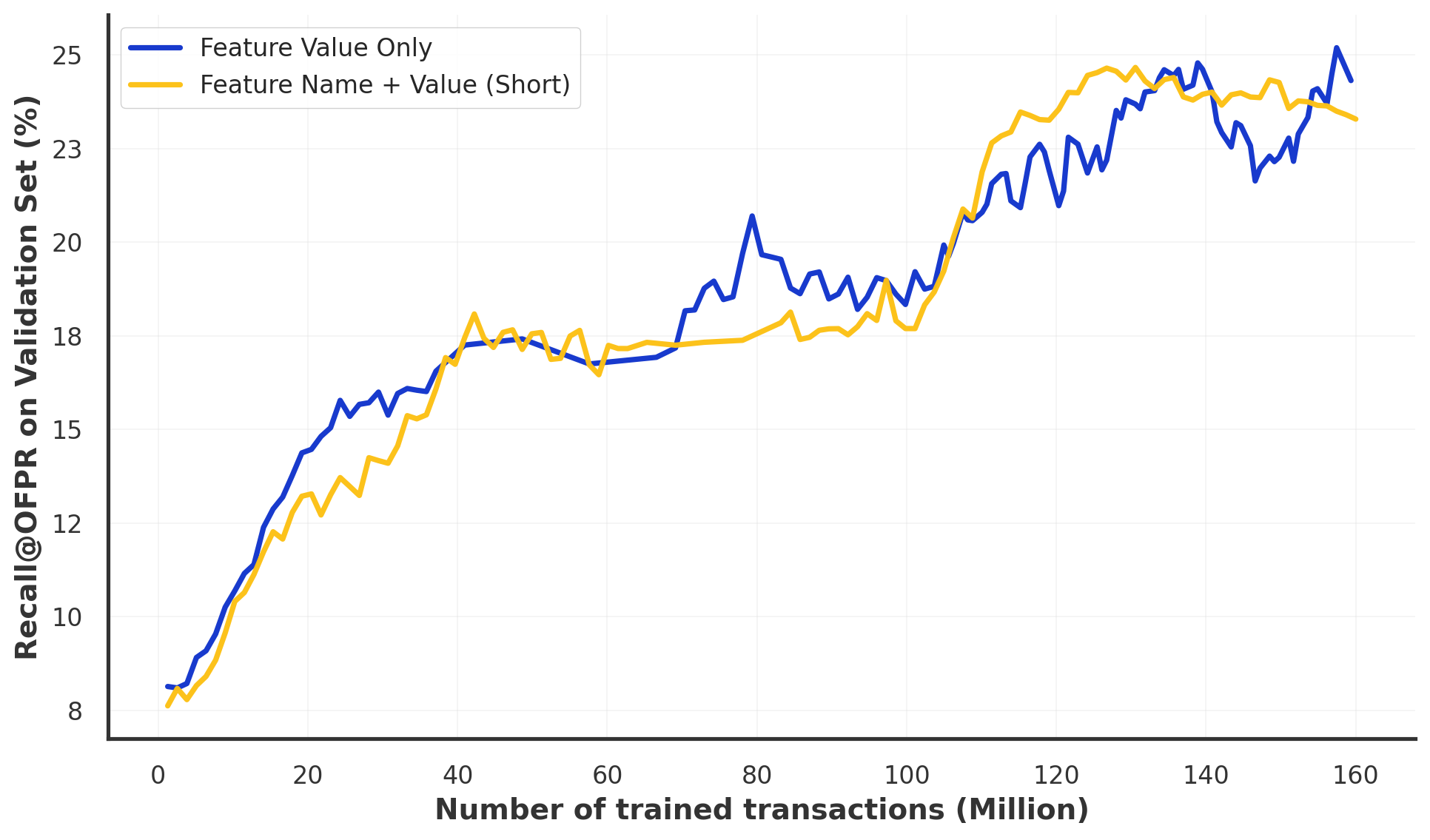}
        \caption{The validation performance of feature value only and feature name value (short) data serialization templates.}
        \label{fig:serialization}
    \end{minipage}
    \hfill
    \begin{minipage}[t]{0.48\linewidth}
        \centering
        \includegraphics[width=\linewidth]{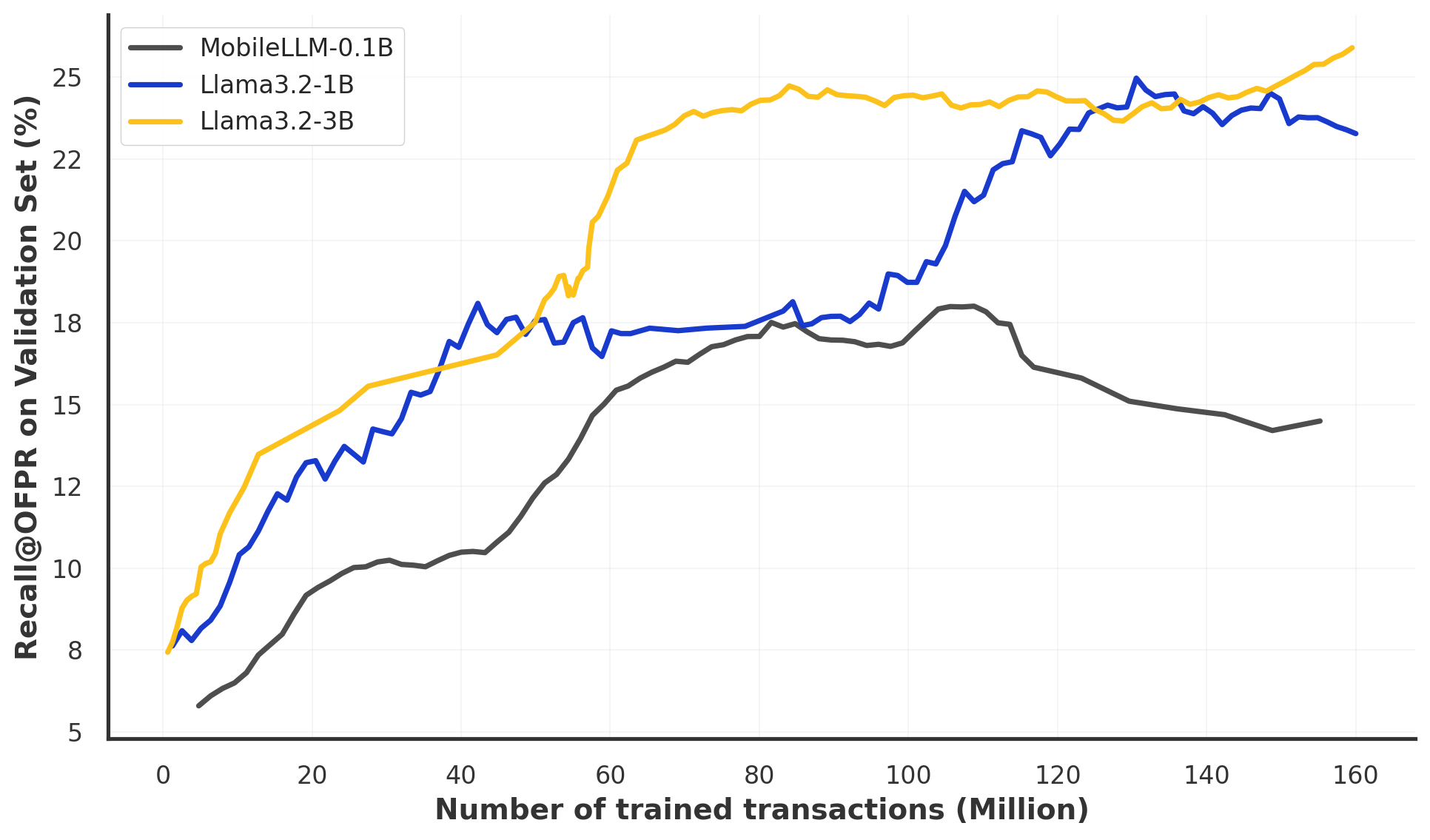}
        \caption{The validation performance of different LLMs on the feature name + value (short) template.}
        \label{fig:diff_llms}
    \end{minipage}
\end{figure}

\begin{figure}
    \centering
   \begin{minipage}[t]{0.48\linewidth}
        \centering
        \includegraphics[width=\linewidth]{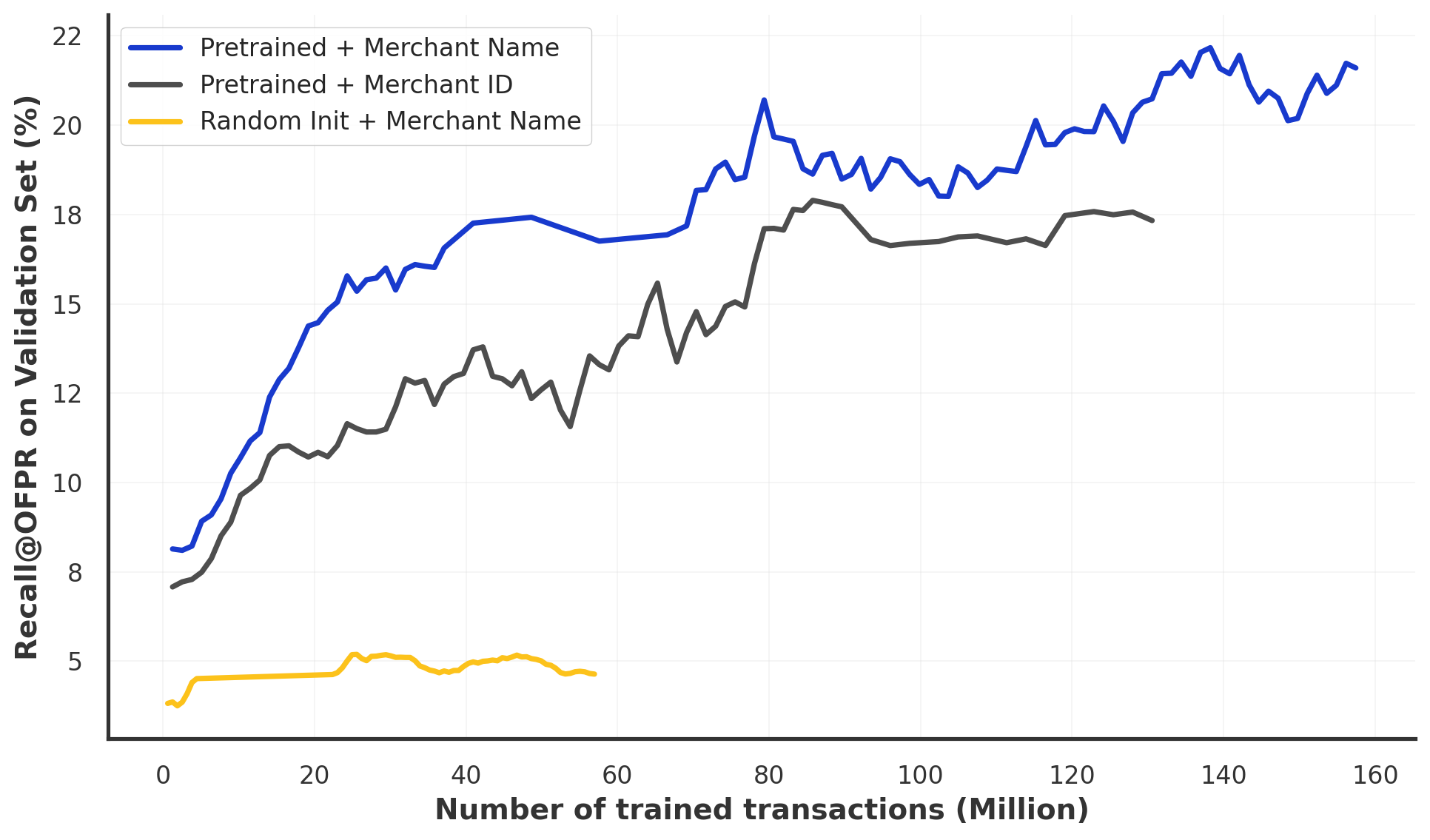}
        \caption{The validation performance of training without pretrained LLM weights and without textual merchant names.}
        \label{fig:pretrain}
    \end{minipage}
    \hfill
    \begin{minipage}[t]{0.48\linewidth}
        \centering
        \includegraphics[width=\linewidth]{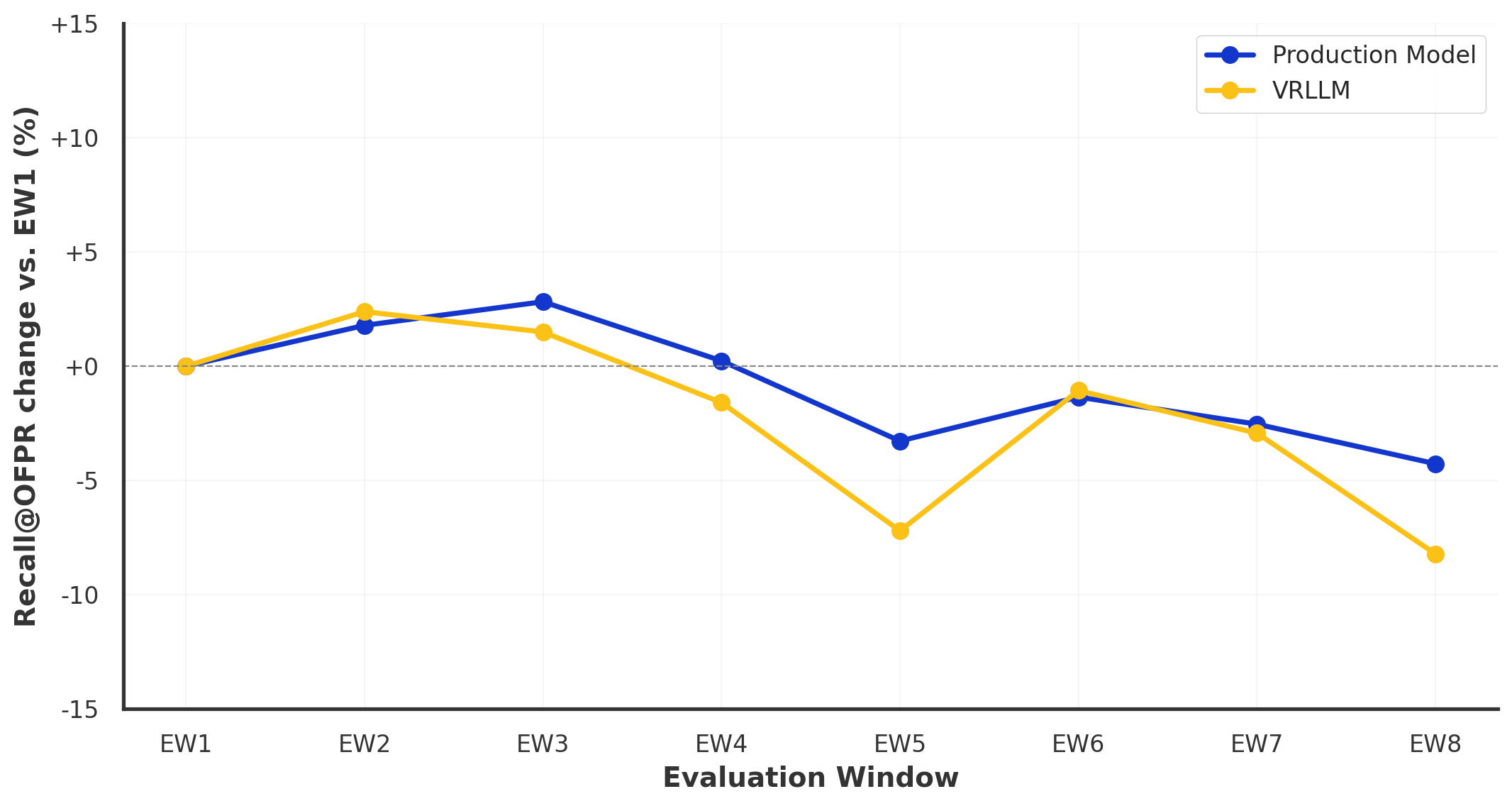}
        \caption{The Recall@OFPR change of VRLLM and production
        model on eight consecutive evaluation time windows.}
        \label{fig:temporal-decay}
    \end{minipage}
\end{figure}

Due to the high computational cost of ablation experiments, we cap the training data at 160M transactions and evaluate on a smaller validation set for the studies in Figs.~\ref{fig:serialization}, \ref{fig:diff_llms}, and \ref{fig:pretrain}. This setup is sufficient to support our conclusions.
All ablation studies use the same hardware, random seed, training recipe, and data, except for the component under investigation. The curves are smoothed across adjacent steps.

\begin{table}[t]
    \centering
    \footnotesize
    \renewcommand{\arraystretch}{1.1}
    \begin{minipage}[t]{0.48\columnwidth}
        \vspace{0pt}
        \centering
        \caption{Validation performance of VRLLM fine-tuned on
        1M transactions with different templates.}
        \label{tab:three_templates}
        \resizebox{\linewidth}{!}{%
            \begin{tabular}{lcc}
                \toprule
                \textbf{Template} & \textbf{Recall@OFPR} & \textbf{PR-AUC} \\
                \midrule
                Feature Value Only & 31.9 & 22.0 \\
                Feature Name + Value (Short) & 33.3 & \textbf{22.6} \\
                Feature Name + Value (Long) & \textbf{35.9} & 22.5 \\
                \bottomrule
            \end{tabular}%
        }
    \end{minipage}
    \hfill
    \begin{minipage}[t]{0.51\columnwidth}
        \vspace{0pt}
        \centering
        \caption{Validation performance of encoder models vs. VRLLM fine-tuned on 1M transactions with short template.}
        \label{tab:embedding-model-ablation}
        \resizebox{\linewidth}{!}{%
            \begin{tabular}{llcc}
                \toprule
                \textbf{Architecture} & \textbf{Model (Size)} & \textbf{Recall@OFPR} & \textbf{PR-AUC} \\
                \midrule
                Encoder & NV-Embed-v2 (8B) & 6.5 & 11.0 \\
                Encoder & BERT (336M)  & 8.2 & 13.0 \\
                Decoder & VRLLM (1B) & \textbf{35.9} & \textbf{22.5} \\
                \bottomrule
            \end{tabular}%
        }
    \end{minipage}
\end{table}

\textbf{Encoder vs. decoder with 1M training examples (Table~\ref{tab:embedding-model-ablation}).}
In the early stage of this project, we used a small dataset (1M transactions) to experiment with different model architectures.
We test whether bidirectional text embedding models are better suited to encoding serialized tabular inputs than a causally masked decoder.
The encoder-style alternatives are BERT~\citep{devlin2019bert} and NV-Embed-v2~\citep{lee2024nvembed}, a more recent embedding model derived from a Mistral~\citep{jiang2023mistral7b} backbone.
Their average-pooled sequence representations are passed to a feed-forward classification head and the entire model is fine-tuned for the transaction classification task.

Under the tested 1M-example configurations, VRLLM outperforms both encoder baselines. Because the models differ in pretraining data/objectives, size, pooling, optimization, and adaptation method as well as attention mask, this experiment does not isolate a general advantage of decoder-only architecture.

\textbf{Three data templates with 1M training examples (Table~\ref{tab:three_templates}).} 
We also experiment with three different serialization templates.
Table~\ref{tab:three_templates} shows that the semantically enriched long template achieves the best Recall@OFPR when only 1M training examples are available.
This suggests that \textit{the semantically rich template is preferable in the low-data regime because its natural-language descriptions provide a useful semantic prior}.
However, as we scale training on more data (see next paragraph), the benefit of semantic enrichment diminishes.
 
\textbf{Value-only vs. short serialization templates (Fig.~\ref{fig:serialization}).}
As we trained on more data, the feature-value-only and feature-name-plus-value (short) templates follow broadly similar learning curves.
The short template has a modest advantage at several intermediate stages, but this advantage narrows with additional data, and the value-only template ultimately performs best at the largest training scale.
This result indicates that explicit feature-name semantics mainly improve data efficiency: \textit{when sufficient labeled data are available, VRLLM can infer predictive regularities directly from the ordered feature values while benefiting from the shorter, more token-efficient input}.
We posit that feature names or markers are still necessary if feature order varies across inputs.

\textbf{LLMs with different sizes (Fig.~\ref{fig:diff_llms}).}
The experiment reveals a clear capacity-dependent learning pattern. Llama 3.2 3B improves more rapidly and maintains the strongest validation performance through most of training, whereas Llama 3.2 1B closes much of the gap after seeing more than 100 million transactions. MobileLLM 0.1B~\citep{liu2024mobilellm} improves initially but plateaus and then declines slightly at the largest data scales. \textit{The 1B backbone offers a favorable balance among classification performance, fine-tuning and inference cost.}

\textbf{Effect of LLM pretraining and input semantics (Fig.~\ref{fig:pretrain}).}
We investigate whether LLM pretraining and semantic information in the tabular data improve classification performance.
We train a Llama 3.2 1B backbone from scratch by reinitializing the attention and FFN layers (the yellow line).
Its performance is much worse than fine-tuning with pretrained weights.
Because the merchant-name field carries the richest semantic information in our transaction data, we map and replace each merchant name with a unique string-based ID consistently across the training and test data to investigate the impact of merchant names on classification performance.
As shown in Fig.~\ref{fig:pretrain}, there is a consistent, substantial gap between training with merchant names and merchant IDs.
We conclude that \textit{the pretrained weights provide a strong starting point for downstream adaptation, and the merchant-name semantics further improve the classifier’s generalization}.
Merchant names offer an extrapolation advantage through their semantic information, in contrast to the out-of-distribution (OOD) generalization limitations of merchant IDs. Please refer to Sec.~\ref{sec:deploy} for more evidence on merchant generalization performance.

\subsection{Deployment-Oriented Validation}
\label{sec:deploy}

\textbf{Temporal performance stability (Fig.~\ref{fig:temporal-decay}).}
To validate the deployment viability of VRLLM, we test its performance alongside that of the production model offline over eight consecutive evaluation time windows without model update.
We cannot disclose the window duration or transaction volume per window due to the company's policy.
Across these windows, VRLLM and the production model exhibit roughly the same temporal trajectory. VRLLM remains competitive throughout and its performance changes closely track those of the production model rather than showing additional model-specific deterioration. This comparable pattern suggests that \textit{VRLLM responds to temporal distribution shift similarly to the deployed reference model}, supporting its suitability for deployment under the same monitoring and model-refresh practices.

\textbf{Generalization to unseen merchants (Table~\ref{tab:merchant_ood_vrllm_prod}).}
A production model continually encounters new and long-tail merchants absent from its training data, so we test whether VRLLM's classification ability depends on seen merchants.
We build an \emph{unseen-merchant} test set by selecting test transactions whose merchant names do not appear in the training data, yielding millions of transactions.
We also construct four seen-merchant control populations, each sampled to the same size and label distribution but matched on one categorical field (merchant category code, product code, dollar bucket, or transaction day of week) to reduce differences unrelated to merchant familiarity.

Table~\ref{tab:merchant_ood_vrllm_prod} reports PR-AUC and Recall@OFPR for VRLLM and the production model on all five populations, each normalized to that model's own Matched-by-Transaction-Day population (100\%).
On the unseen-merchant population, VRLLM shows a real downgrade (86.7\% PR-AUC, 81.5\% Recall@OFPR), confirming some dependence on merchant memorization, but not a collapse in performance. The production model's downgrade on the same population is larger on both metrics (73.8\% PR-AUC, 68.0\% Recall@OFPR), while the two models track each other closely across the four seen-merchant controls.
\textit{This indicates VRLLM generalizes to unseen merchants somewhat better than the production model, and that both metrics remain within a manageable range supporting VRLLM's viability when handling the steady stream of new and long-tail merchants in production.}

\begin{table}[tbh]
    \centering
    \caption{VRLLM vs.\ the production model on the unseen-merchant test set and four seen-merchant control populations. PR-AUC and Recall@OFPR are each normalized to that
    model's own Matched-by-Transaction-Day population (100\%) so the relative drop
    for each population is directly comparable.}
    \label{tab:merchant_ood_vrllm_prod}
    \footnotesize
    \begin{tabular}{lcccc}
            \toprule
            & \multicolumn{2}{c}{\textbf{Relative PR-AUC (\%)}} & \multicolumn{2}{c}{\textbf{Relative Recall@OFPR (\%)}} \\
            \cmidrule(lr){2-3} \cmidrule(lr){4-5}
            \textbf{Population} & \textbf{VRLLM} & \makecell{\textbf{Production} \\ \textbf{Model}} & \textbf{VRLLM} & \makecell{\textbf{Production} \\ \textbf{Model}} \\
            \midrule
            Matched by Transaction Day & \textbf{100.0} & \textbf{100.0} & \textbf{100.0} & \textbf{100.0} \\
            Matched by Product Code    & 99.4 & 99.9 & 97.8 & 97.4 \\
            Matched by Dollar Bucket   & 98.8 & 99.1 & 93.5 & 93.8 \\
            Matched by MCC             & 92.1 & 86.7 & 87.9 & 83.7 \\
            Unseen Merchant            & 86.7 & 73.8 & 81.5 & 68.0 \\
            \bottomrule
    \end{tabular}
\end{table}

\begin{table}[tbh]
    \centering
    \caption{Disagreement analysis of VRLLM and production model on the same evaluation data. The number under each proprietary feature (A, B, and C) represents its average value in the corresponding true-positive group.}
    \label{tab:disagree}
    \resizebox{0.65\linewidth}{!}{
        \begin{tabular}{lcccc}
            \toprule
            \multicolumn{1}{l}{\textbf{Model}}
                & \makecell{\textbf{Unique true positives} \\ \textbf{(\% of all positives)}}  
                & \textbf{Feature A} & \textbf{Feature B} & \textbf{Feature C} \\
            \midrule
            VRLLM            & 8.0\% & 53.2 & 8.7 & 5.5 \\
            Production Model & 7.9\% & 33.6 & 2.7 & 13.1 \\
            \bottomrule
        \end{tabular}
    }
\end{table}

\textbf{Disagreement and ensemble analysis (Table~\ref{tab:disagree}).}
Since VRLLM and the production model are trained on the same set of features, we want to understand their prediction differences and the potential for ensembling.
Following our prior work on learning from disagreement~\citep{wang2022learning}, we carry out a post-hoc analysis of the predictions from both models on the same evaluation data.
At the comparable operating threshold, VRLLM uniquely captures 8.0\% of all true positive transactions, while the production model uniquely captures 7.9\%.

The feature-level analysis in Table~\ref{tab:disagree} reveals that the unique positives captured by each model have distinct average values across selected features.
Moreover, as shown in Table~\ref{tab:main_results}, averaging the prediction scores of VRLLM and the production model yields improvements of 7.6 and 3.4 points in Recall@OFPR and PR-AUC, respectively.
These results collectively demonstrate that \textit{VRLLM learns predictive patterns distinct from those of the production model despite being trained on the same features}.
These complementary patterns improve classification performance through ensembling, which is a viable way to put the model in production.

\textbf{Note on model robustness.}
Since our classification task is sensitive to numerical precision and classification thresholds, we post-process and calibrate the LLM’s predicted probabilities during inference to maintain robust deployment performance.
We do not use the post-training or temperature-based calibration techniques employed by Jev-like decision models, since we use our own business logic to calibrate probabilities.
We also find that fixing the input format, padding settings, GPU kernels, and software and hardware configurations is critical to maintaining deterministic predictions and robust performance during inference.

\section{Model Interpretation}
\label{sec:interpretation}

Sec.~\ref{sec:experiments} shows that VRLLM performs well; this section asks why. Understanding the underlying mechanisms matters for two practical reasons: first, it helps us distinguish generalizable predictive signals from narrow memorized shortcuts, informing trust in a decision-critical production model; second, it surfaces hypotheses that can guide future model design.
We examine VRLLM from two angles: a \emph{data perspective} (Sec.~\ref{sec:data_interpretation}), where counterfactual feature swaps reveal what prior each input field contributes to the prediction, and a \emph{model perspective} (Sec.~\ref{sec:model_interpretation}), where causal ablation of individual attention heads and FFN layers identifies which internal components the prediction depends on.

\subsection{Data Perspective}
\label{sec:data_interpretation}

To isolate the prior associated with an individual feature, we replace only that feature's serialized value in holdout transactions, keep every other input token fixed, and measure the resulting change in the predicted positive-class probability~\citep{lundberg2017unified}. We aggregate these changes for each source--target pair and examine the two ground-truth classes separately. This intervention measures the model's sensitivity to a feature value and the prior carried by its text representation; because the swap can break correlations among fields and push the input off the data manifold, it should not be interpreted as a causal effect of changing the underlying transaction~\citep{aas2021explaining}. Tables~\ref{tab:product-code-prior} and~\ref{tab:dollar-bucket-prior} use pseudonymous values to summarize the two distinct response patterns without disclosing proprietary feature values.

\textbf{Product code: VRLLM looks up each code's historical rate.}
The product code is an internally defined field, provided to VRLLM as a categorical feature.
The product-code swaps (Table~\ref{tab:product-code-prior}) show VRLLM behaving like a lookup table over code identities: moving from a code with a lower positive rate in the training data to one with a higher rate raises the model's predicted probability, and the reverse lowers it, even though every other input feature stays unchanged.
This happens in the same direction for both ground-truth classes, so the code is not applying a class-specific rule; it is adding a fixed offset based on how often that code was positive in training, i.e., a base rate learned directly from training data statistics.
Because the codes are nominal (there is no natural low-to-high order among them), this is best explained as a stored, per-code lookup value, not as reasoning over the code's value. See Appendix~\ref{appdix:interpretation_product_id} for the underlying counterfactual-swap results supporting this pattern.

\begin{table}[tbh]
    \centering
    \caption{A summary of the prior learned for a product-code feature. Code names and prior levels are illustrative for compliance; the response directions summarize the feature-swap analysis.}
    \label{tab:product-code-prior}
    \footnotesize
    \renewcommand{\arraystretch}{1.12}
    \begin{tabularx}{0.83\linewidth}{
        @{}
        >{\raggedright\arraybackslash}p{0.08\linewidth}
        >{\raggedright\arraybackslash}p{0.08\linewidth}
        >{\raggedright\arraybackslash}p{0.31\linewidth}
        >{\raggedright\arraybackslash}X
        @{}
    }
        \toprule
        \textbf{Source} & \textbf{Target} & \textbf{Target association} & \textbf{Typical prediction probability response} \\
        \midrule
        Code-A & Code-B & Much higher training-sample positive rate & Increases for both ground-truth classes \\
        Code-B & Code-A & Much lower training-sample positive rate & Decreases for both ground-truth classes \\
        \bottomrule
    \end{tabularx}
\end{table}

\textbf{Dollar bucket: VRLLM associates a higher bucket with a higher positive probability, regardless of its historical rate.}
The dollar bucket is a bucketed transaction-amount range, not the exact dollar amount, given to VRLLM as a categorical feature. Dollar-bucket swaps (Table~\ref{tab:dollar-bucket-prior}) behave differently from product-code swaps: moving from a low bucket to a high bucket raises the predicted probability, and the reverse lowers it, even though the buckets have similar positive rates in the training data.
So this is not VRLLM recalling a per-bucket historical rate the way it does for product code (there is none to recall--the rates are nearly identical across buckets). Instead, VRLLM treats the bucket's position in the low-to-high ordering itself as the signal, applying the rule ``a higher bucket number means higher positive probability'' uniformly across buckets. Together with our frontier model analysis in Sec.~\ref{sec:zeroshot_exp}, these findings suggest that this prior may originate from pretraining and remain largely unchanged after fine-tuning. See Appendix~\ref{appdix:interpretation_dollar_bucket} for the underlying counterfactual-swap results supporting this pattern.

\begin{table}[tbh]
    \centering
    \caption{A summary of the prior learned for the dollar-bucket feature. Buckets are described only by relative direction (low to high) to avoid revealing production boundaries.}
    \label{tab:dollar-bucket-prior}
    \footnotesize
    \renewcommand{\arraystretch}{1.12}
    \begin{tabularx}{0.85\linewidth}{
        @{}
        >{\raggedright\arraybackslash}p{0.11\linewidth}
        >{\raggedright\arraybackslash}p{0.11\linewidth}
        >{\raggedright\arraybackslash}p{0.25\linewidth}
        >{\raggedright\arraybackslash}X
        @{}
    }
        \toprule
        \textbf{Source} & \textbf{Target} & \textbf{Training data positive rate} & \textbf{Typical prediction probability response}  \\
        \midrule
        Low bucket & High bucket & Similar across buckets & Increases for both ground-truth classes \\
        High bucket & Low bucket & Similar across buckets & Decreases for both ground-truth classes \\
        \bottomrule
    \end{tabularx}
\end{table}

\textbf{Different features induce different priors.}
Taken together, the counterfactual swaps on two representative features show that VRLLM does not apply a uniform rule to every categorical feature.
The product code primarily supplies an identity-specific empirical base rate, while the dollar bucket supplies an ordered magnitude prior.
\textit{The final prediction score of VRLLM is based on the combination of priors induced by various input features}.

\subsection{Model Perspective}
\label{sec:model_interpretation}

Having characterized sensitivity to two input features, we now turn from \emph{what} the model uses to \emph{how} it computes it. We treat each attention head and each FFN sublayer as an individually testable component (a \emph{circuit})~\citep{elhage2021mathematical} of the residual stream, and intervene on one circuit at a time by replacing its activation with a class-agnostic mean-ablation value while holding every other circuit fixed. This is a causal instead of correlational analysis~\citep{meng2022locating,wang2023interpretability}. The positive class signal is \emph{linearly decodable} from a layer if a linear classifier trained on that layer's residual stream alone can predict the label above chance~\citep{alain2016understanding}; observing this only shows association. If instead we ablate a circuit and that decodability disappears, the circuit is \emph{necessary} for the positive class signal.

\begin{figure}
    \centering
    \includegraphics[width=0.6\linewidth]{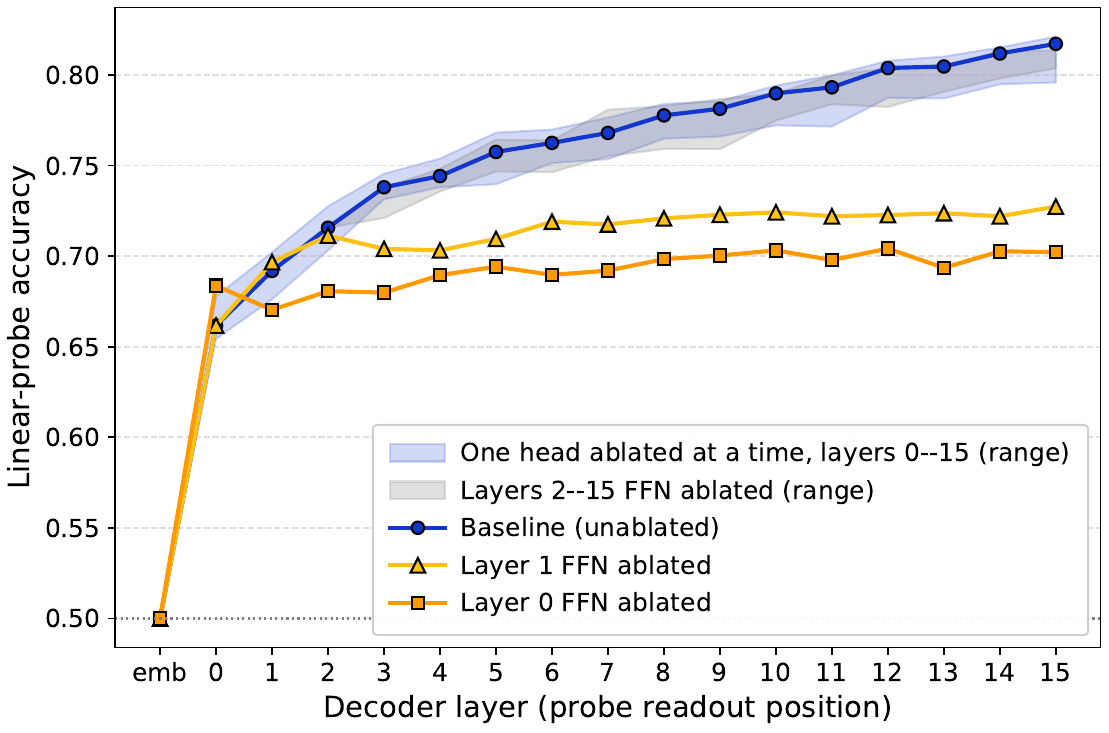}
    \caption{Balanced binary linear-probe accuracy by decoder layer under the baseline (unablated) model, with layer-0 and layer-1 FFN ablation shown individually and the range spanning single-head ablation (layers 0--15, $512=32{\times}16$ heads in total) and layer-2--15 FFN ablation shown as bands.}
    \label{fig:probe_ablated_early_mlp}
\end{figure}

On a holdout evaluation set of 20K transactions (10K positive, 10K negative), we train a binary linear probe to predict the transaction label from the residual stream after each decoder layer. Under normal operation, the probe's discriminative accuracy climbs steadily with depth; ablating either of the earliest two FFN layers (layer 0 or layer 1) causes that climb to plateau early, i.e., within two to three layers after the ablation point, well below the baseline's final-layer accuracy.
Ablating any other single circuit leaves the climb essentially intact (Fig.~\ref{fig:probe_ablated_early_mlp}). The earliest two FFN layers are therefore critical for the positive class signal the rest of the network relies on. No other individual component produced a comparably large effect. 

A plausible architectural explanation is the \emph{curse of depth}~\citep{sun2025curse}: VRLLM's backbone uses pre-normalization, where each layer adds its output directly onto a residual stream that every later layer also reads from and writes to. Because this stream's scale grows with depth, by the later layers each new layer's own contribution is small relative to what has already accumulated, so late layers end up barely modifying the representation they receive rather than substantially transforming it. Early layers do not have this handicap, so they are where the network has the most room to shape the representation. This independently predicts that early layers should matter disproportionately, consistent with our ablation results.

This explanation accounts for the FFN result, but not for why no single early attention head shows the same effect: curse of depth is about \emph{depth} (early vs. late) and says nothing about redundancy \emph{within} a layer.
Llama 3.2 1B’s multi-head attention has 32 parallel heads at each layer, so one head's output can likely be compensated by the other 31; the FFN sublayer has no such parallel redundancy. We do not have an architectural explanation for the head result and leave it as an open question.

Mean-ablation only tells us a circuit is necessary. It cannot tell us what the circuit actually does to the prediction, because replacing its output with an average value erases any directional information. Activation steering~\citep{turner2023activation} and activation patching~\citep{meng2022locating} intervene with a specific value instead, e.g., the opposite class's typical activation, or another example's activation.
Therefore they could show whether the earliest FFN layers push predictions toward the right class or simply gate the signal on and off. We leave this to future work.

\section{Related Work}
\label{sec:related_work}

\subsection{Transaction Representation Learning}

Transaction representation learning encompasses graph-based, sequential, and record-level approaches.
Graph neural networks (GNNs) capture dependencies among interacting entities, e.g., banking transaction graphs combine structural and edge-level information~\citep{minakawa2022graph}. Sequential methods instead encode behavioral histories: PDT~\citep{dai2023pdt} learns user and merchant embeddings by encoding behavior sequences from different views with Transformers, while FATA-Trans~\citep{zhang2023fata} uses Transformers to capture within-record feature interactions and temporal dependencies.
Payment foundation models further explore purchase-sequence pretraining and downstream task adoption~\citep{skalski2023towards, dou2025transactiongpt, yeh2026treasure, ostroukhov2026pragma}. These methods generally consume graphs or sequences instead of independent serialized rows and rely on dedicated graph or sequence encoders.

For individual records, modern tabular architectures introduce explicit feature-interaction and attention-based inductive biases. DCNv2 combines an enhanced cross network with deep neural layers to learn explicit and implicit feature interactions~\citep{wang2021dcnv2}. TabTransformer~\citep{huang2020tabtransformer} contextualizes categorical fields, while SAINT~\citep{somepalli2021saint} combines feature-wise and intersample attention with contrastive pretraining. Nevertheless, boosted trees~\citep{chen2016xgboost, ke2017lightgbm} remain competitive on typical tabular benchmarks, with relative performance depending on dataset characteristics and tuning budgets~\citep{grinsztajn2022tree,mcelfresh2023neural}.

Recent tabular foundation models combine pretraining with in-context learning to transfer predictive capabilities across datasets. TabPFN-3~\citep{grinsztajn2026tabpfn3} scales to million-row tables, while TabICLv2~\citep{jingangtabicl} improves large-table generalization with a scalable-softmax attention.
TabFM~\citep{kong2026tabfm} and KumoTabular~\citep{nvidia2026kumotabular} support in-context classification and regression using causal-model-generated tables and structured attention mechanisms.
Together, these models enable transferable, dataset-level inference without training a separate predictor from scratch for each task.
However, the in-context fitting/inference can still be computationally intensive.

VRLLM follows a different approach: it serializes transaction records and fine-tunes a general pretrained LLM on billions of transactions.
Rather than introducing a domain-specific backbone or learning a separate embedding table for each categorical field, it reuses a shared pretrained language-model vocabulary to represent extremely high-cardinality features.
Its distinction is therefore the combination of transaction serialization, vocabulary and pretrained knowledge reuse, and large-scale domain supervision.

\subsection{LLM for Transaction Data}

LLM-based approaches to tabular and transaction data differ primarily in how structured inputs are represented and how pretrained models are adapted~\citep{ruan2024language}. Text serialization provides a direct interface to existing tokenizers and language-model representations: TabLLM~\citep{hegselmann2023tabllm} serializes table rows for prompting or few-shot fine-tuning, while LLM4ES~\citep{shestov2025llm4es} extends this approach to event sequences by fine-tuning on enriched textual descriptions to learn user embeddings.
Other methods introduce specialized tokenization or embedding alignment to accommodate structured behavioral inputs. SeqLLM~\citep{li2026seqllm} combines dedicated event tokens and a lightweight projector with supervised fine-tuning, whereas LATTE~\citep{fadeev2025latte} and MINT~\citep{kamalaruban2026mint} connect transaction encoders to language representations through contrastive alignment and projected embedding injection, respectively.
LLMs can also serve as frozen feature extractors, generating embeddings of textualized rows for downstream tabular models~\citep{koloski2025llm, fan2025enhancing}.
Together, these approaches span direct textual adaptation, specialized input representations, and embedding-based transfer.

Complementing work on LLM tabular data encoding, recent work explores decision-oriented prediction instead of free-form generation.
Jev~\citep{almeida_jev_2026} and the OpenAI Decisions API~\citep{openai_decisions_api_2026} return typed predictions, probabilities, or scores that applications can consume directly.
An open-weight model such as Perplexity Decider~\citep{pplx_decider_v1_2_2026} adapts a pretrained backbone~\citep{qwen38} for structured decision tasks. These systems share the goal of avoiding lengthy LLM responses when the desired output is a bounded decision, although their training procedures and prediction interfaces differ.

VRLLM shares this decision-oriented objective while following TabLLM, LLM4ES, and SeqLLM in adapting pretrained language models through task-specific supervision on serialized tabular data.
VRLLM retains the tokenizer and pretrained vocab-embedding instead of introducing a dedicated event vocabulary, field-specific embedding tables, or an external transaction encoder.
This choice avoids vocabulary expansion and additional representation-alignment stages, allowing high-cardinality field values to be represented compositionally through shared tokens.

\subsection{Predictive Model and Transformer Interpretability}
Interpretability methods for predictive models and Transformers span a spectrum: input-level attribution asks what a feature contributes to a prediction, component-level analysis asks which internal parts compute it, and probing asks what information a representation carries. We organize this subsection along that spectrum.

Input-level feature attribution typically perturbs or removes a feature and measures the resulting change in model output, formalized through a game-theoretic allocation of credit across features~\citep{lundberg2017unified}. Because this requires evaluating the model on feature combinations absent from the training distribution, attribution can be distorted when features are correlated, motivating conditional or dependence-aware approximations~\citep{aas2021explaining}. Our counterfactual feature-value swaps follow this input-perturbation tradition. However, instead of a numerical attribution score, we characterize the qualitative response pattern each field induces, e.g., base-rate lookup for product code and ordinal generalization for dollar bucket.

Mechanistic interpretability treats attention heads and FFN layers as individually testable \emph{circuits} wired to the residual stream to contribute additional signal~\citep{elhage2021mathematical}. Prior work localizes circuit-level computation by patching or restoring hidden states~\citep{meng2022locating,wang2023interpretability}, or by steering activations along a learned direction~\citep{turner2023activation}. We pair this causal logic with linear probing of anomaly signals~\citep{alain2016understanding,hewitt2019designing,belinkov2022probing}. A probe shows what information a representation carries, but not that the model actually uses it. To test actual use, we ablate individual circuits and probe the anomaly signal in layer-wise activations before and after ablation (Fig.~\ref{fig:probe_ablated_early_mlp}). This way, we can distinguish a signal that is merely present from one on which the model actually depends on it. A separate line of work instead prunes or skips whole layers and finds that many contribute little to final performance~\citep{gromov2024unreasonable,men2024shortgpt}, broadly consistent with our finding that the earliest FFN layers are disproportionately important relative to the other circuits we test (Figs.~\ref{fig:probe_ablated_early_mlp}).

\section{Conclusion}
\label{sec:conclusion}

This paper demonstrates that a pretrained decoder-only LLM can be adapted into a competitive industrial-scale tabular classifier by serializing transaction records and fine-tuning on domain-specific labels.
Trained on billions of records, VRLLM achieves the strongest performance among individual models in our benchmark.
Our ablation study and interpretability analysis provide further insights into its performance and predictive behavior.
Although our findings are limited to a single proprietary task, they offer a practical recipe for LLM-based tabular prediction and a testable interpretation of how these models make predictions.

\bibliographystyle{unsrtnat}
\bibliography{refs}

\clearpage
\beginappendix
\section{Interpretation: Product Code Swap Analysis}
\label{appdix:interpretation_product_id}

Sec.~\ref{sec:data_interpretation} argues that VRLLM treats the product code as a lookup table of training-data rates instead of reasoning over the code's value. This appendix substantiates that claim with the full counterfactual-swap results.

\begin{figure}[t]
    \centering
    \includegraphics[width=\linewidth]{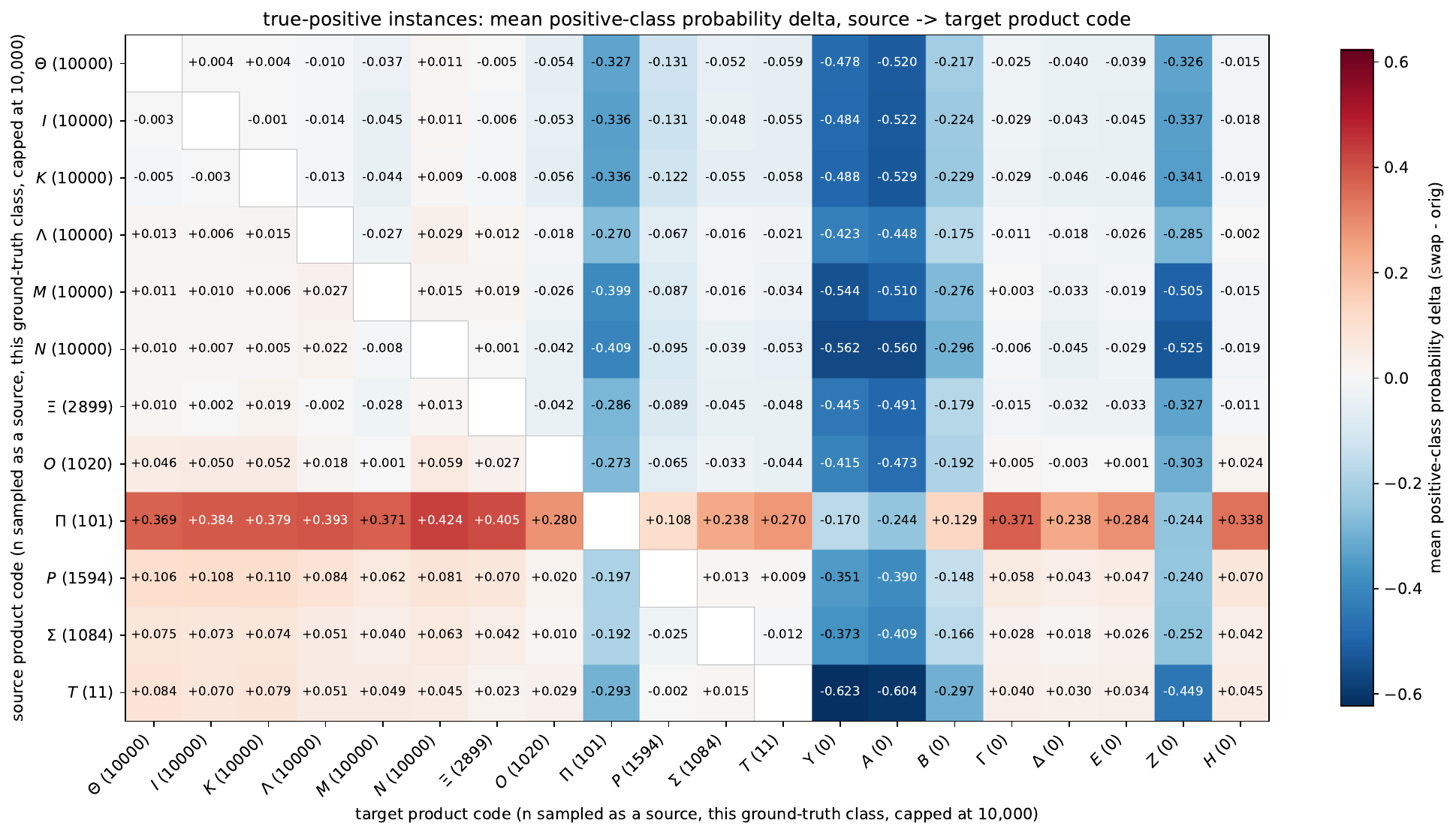}
    \caption{Product-code counterfactual swap, true-positive source instances: mean signed change in predicted positive-class probability (post-swap minus original) when the source product code (rows) is replaced by the target product code (columns), restricted to a 12-code (rows) by 20-code (columns) subset; rows are a subset of the codes in Table~\ref{tab:product-code-stats-subset}, and columns additionally include code $H$ (never seen during training, absent from Table~\ref{tab:product-code-stats-subset}). Axis labels give, in parentheses, the number of true-positive transactions actually sampled for each code as a source (up to $10{,}000$). Diverging color scale centered at zero.}
    \label{fig:product-code-swap-positive}
\end{figure}

\begin{figure}[t]
    \centering
    \includegraphics[width=\linewidth]{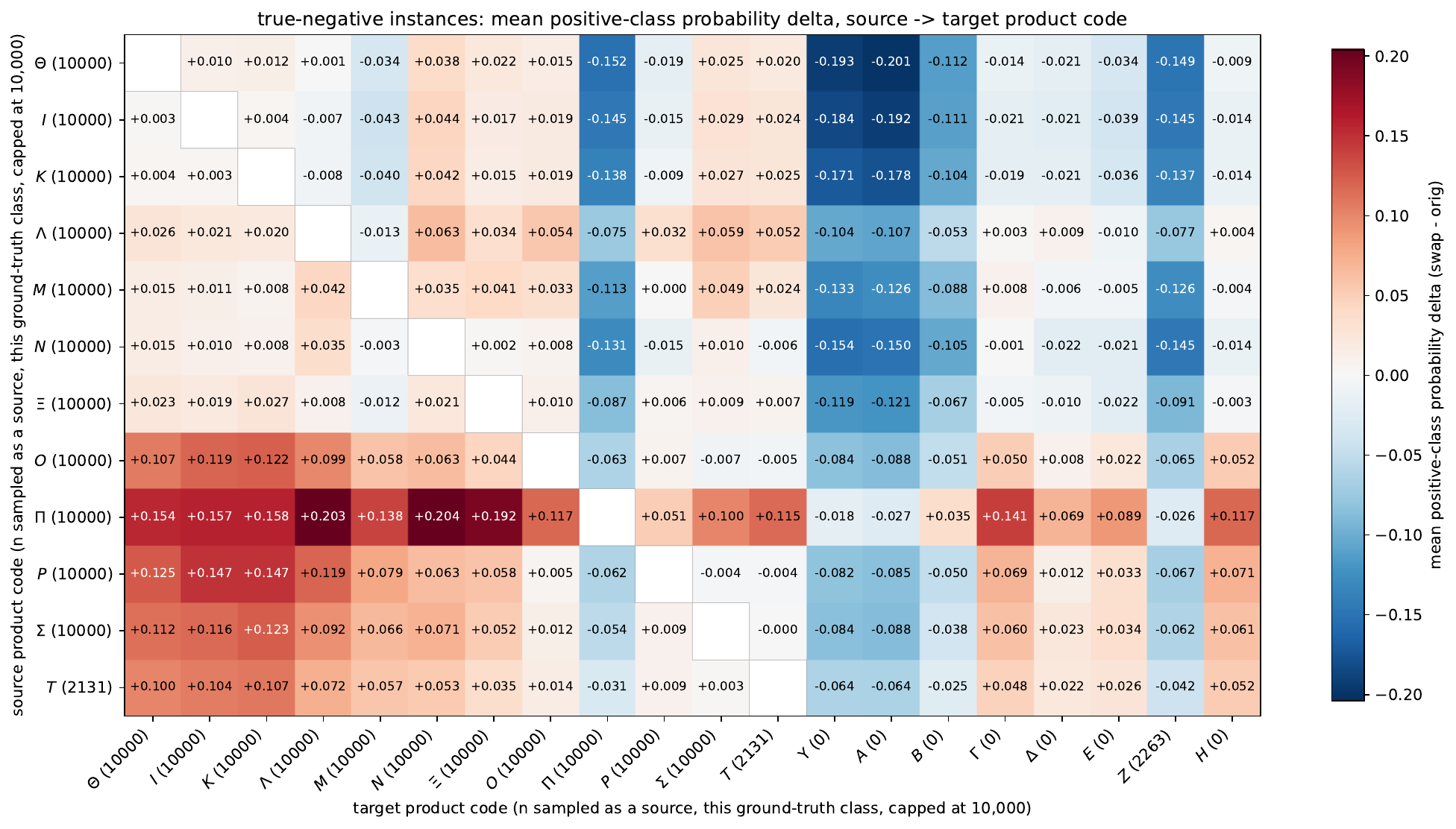}
    \caption{Same construction as Figure~\ref{fig:product-code-swap-positive}, restricted to true-negative source instances.}
    \label{fig:product-code-swap-negative}
\end{figure}

\textbf{Experiment setup.}
Table~\ref{tab:product-code-stats-subset} and Figs.~\ref{fig:product-code-swap-positive}--\ref{fig:product-code-swap-negative} support this pattern with the underlying counterfactual-swap results. Since the product code itself cannot be disclosed, every code is encoded as a capitalized Greek/Latin letter ($\Theta$, $I$, $K$, \ldots); the encoding is arbitrary but fixed across Table~\ref{tab:product-code-stats-subset} and Fig.~\ref{fig:product-code-swap-positive}--\ref{fig:product-code-swap-negative}. Table~\ref{tab:product-code-stats-subset} reports training-set statistics for a 19-code subset, and Figs.~\ref{fig:product-code-swap-positive}--\ref{fig:product-code-swap-negative} report the counterfactual swap matrix restricted to a 12-code (rows) by 20-code (columns) subset for readability; the swap matrix additionally includes code $H$, a placeholder value never seen during training and therefore absent from Table~\ref{tab:product-code-stats-subset}.

\textbf{Sampling method.}
For each source code within a ground-truth class, we sample up to $10{,}000$ held-out test transactions.
We use every available transaction if fewer than $10{,}000$ exist for that (source code, ground-truth class) combination and replace the serialized product-code value with every target code, measuring the resulting change in the predicted positive-class probability.
The actual number of transactions sampled per source code is given in parentheses on the row and column axis labels of Figs.~\ref{fig:product-code-swap-positive}--\ref{fig:product-code-swap-negative}.

\textbf{Result.}
Code $\Pi$ has a near-zero training-set positive rate ($0.24\%$); $\Upsilon$ and $A$, included only as swap targets in Figs.~\ref{fig:product-code-swap-positive}--\ref{fig:product-code-swap-negative}, were never observed as positive in training.
All three act as strong negative-class signals regardless of the swapped instance's true label.
Swapping \emph{away} from $\Pi$ into the high-base-rate $N$ code ($33\%$ training positive rate) produces the largest increase among true-positive instances in this subset ($\Pi\!\to\!N$: $0.412 \to 0.837$, $+0.424$, $n{=}101$), while swapping true-positive instances originally tagged $N$ \emph{into} $\Upsilon$ produces the largest decrease ($N\!\to\!\Upsilon$: $0.753 \to 0.191$, $-0.562$, $n{=}10{,}000$).
The same pattern holds, with a smaller magnitude, for true-negative instances ($\Pi\!\to\!N$: $0.036 \to 0.240$, $+0.204$, $n{=}10{,}000$; $N\!\to\!\Upsilon$: $0.167 \to 0.012$, $-0.154$, $n{=}10{,}000$).

In short, VRLLM appears to treat the product code largely as a base-rate lookup: swapping to a historically low-positive-rate code suppresses the predicted positive-class probability regardless of the transaction's other features and true label, and vice versa for high-positive-rate codes.

\begin{table}[tbh]
    \centering
    \caption{Training-set positive rate for a 19-code subset of the product-code feature (sorted by positive rate, descending). Each code is replaced by a fixed pseudonymous symbol since the underlying product-code values cannot be disclosed; the encoding is arbitrary but matches Figures~\ref{fig:product-code-swap-positive}--\ref{fig:product-code-swap-negative}. Negative-class records were downsampled so these rates are not representative of the true population rate.}
    \label{tab:product-code-stats-subset}
    \footnotesize
    \renewcommand{\arraystretch}{1.12}
    \resizebox{0.9\linewidth}{!}{%
    \begin{tabular}{l*{10}{r}}
        \toprule
        \textbf{Product Code} & $I$ & $K$ & $\Theta$ & $\Xi$ & $E$ & $N$ & $\Lambda$ & $M$ & $O$ & $\Sigma$ \\
        \textbf{Positive Rate} & 45.20\% & 42.67\% & 38.29\% & 35.56\% & 34.23\% & 33.27\% & 30.57\% & 26.71\% & 26.25\% & 24.99\% \\
        \midrule
        \textbf{Product Code} & $P$ & $T$ & $Z$ & $\Pi$ & $\Delta$ & $\Gamma$ & $B$ & $\Upsilon$ & $A$ & \\
        \textbf{Positive Rate} & 16.70\% & 15.76\% & 0.50\% & 0.24\% & 0.00\% & 0.00\% & 0.00\% & 0.00\% & 0.00\% & \\
        \bottomrule
    \end{tabular}%
    }
\end{table}

\section{Interpretation: Dollar Bucket Swap Analysis}
\label{appdix:interpretation_dollar_bucket}

A parallel analysis of the dollar-bucket feature, summarized in Sec.~\ref{sec:data_interpretation}, finds that VRLLM treats a higher bucket as riskier irrespective of its training-data rate. The swap results below support this reading.

\begin{figure}[t]
    \centering
    \includegraphics[width=\linewidth]{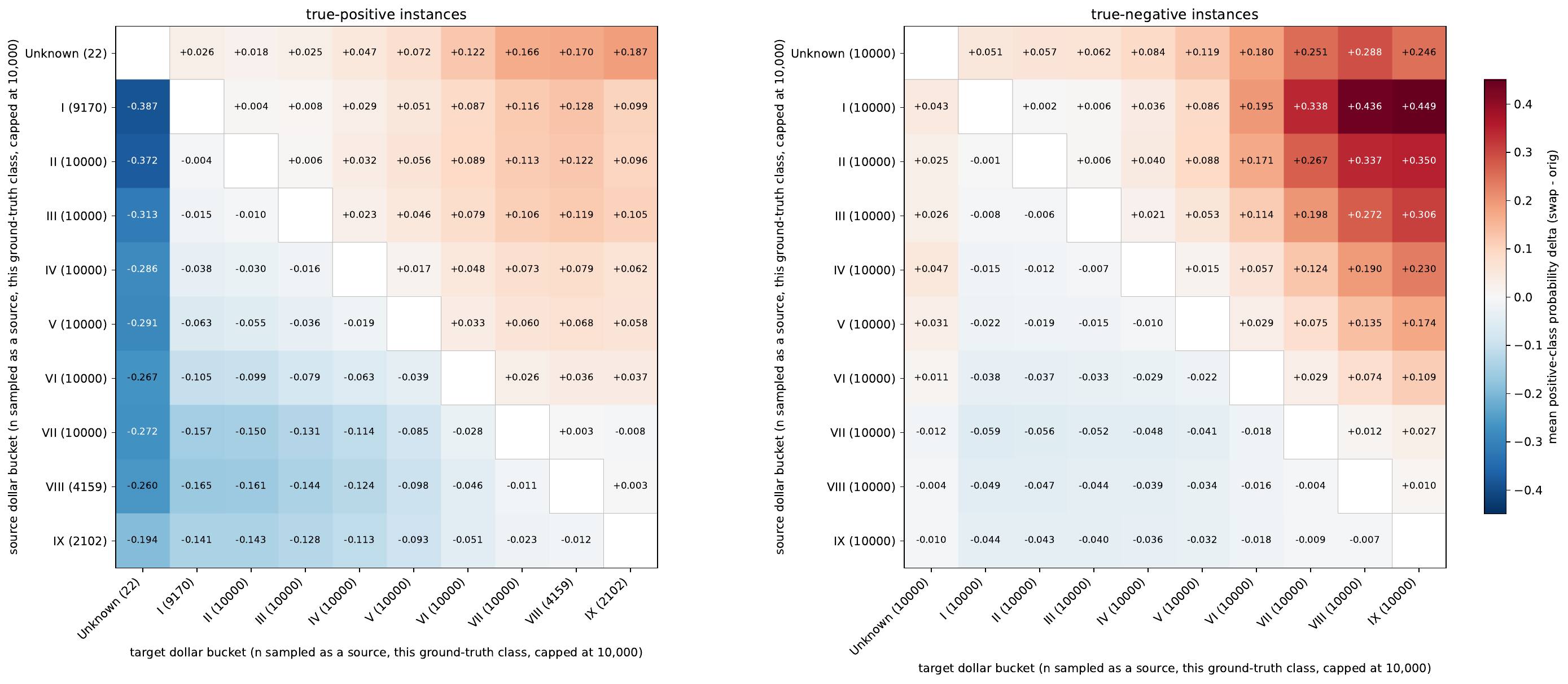}
    \caption{Dollar-bucket counterfactual swap: mean signed change in predicted positive-class probability (post-swap minus original) when the source bucket (rows) is replaced by the target bucket (columns), restricted to ten observed training categories of Table~\ref{tab:dollar-bucket-stats-subset}, arranged in their true increasing order. Left: true-positive source instances; right: true-negative source instances, sharing the same diverging color scale (centered at zero) for direct comparison. Axis labels give, in parentheses, the number of transactions actually sampled for each bucket as a source, for that panel's ground-truth class (up to $10{,}000$).}
    \label{fig:dollar-bucket-swap}
\end{figure}

\textbf{Experiment setup.}
Table~\ref{tab:dollar-bucket-stats-subset} and Fig.~\ref{fig:dollar-bucket-swap} support this pattern with the underlying counterfactual-swap results, restricted to ten categories observed during training for readability.
Since the actual dollar-amount ranges cannot be disclosed, buckets are encoded as Roman numerals in their true increasing order ($\mathrm{I} < \mathrm{II} < \cdots < \mathrm{IX}$); bucket $0$, a non-numeric \emph{unknown} category rather than a dollar amount, is labeled ``Unknown'' and falls outside this ordering.
The encoding is arbitrary only in that it hides the real dollar thresholds and the relative order it displays is the true one.
The data sampling method of this experiment is the same as that used in the product-code swap analysis.

\textbf{Result.}
The most extreme swap, from the lowest bucket $\mathrm{I}$ to the highest bucket $\mathrm{IX}$, raises the predicted positive-class probability for both true-positive instances ($\mathrm{I}\!\to\!\mathrm{IX}$: $0.764 \to 0.863$, $+0.099$, $n{=}9{,}170$) and, more sharply, true-negative instances ($\mathrm{I}\!\to\!\mathrm{IX}$: $0.157 \to 0.606$, $+0.449$, $n{=}10{,}000$); the reverse swap lowers it in both cases ($\mathrm{IX}\!\to\!\mathrm{I}$: $0.674 \to 0.533$, $-0.141$, $n{=}2{,}102$ for true-positive; $0.170 \to 0.126$, $-0.044$, $n{=}10{,}000$ for true-negative).

To sum up, VRLLM appears to treat the dollar bucket as an ordinal risk signal instead of a base-rate lookup: moving a transaction to a higher bucket raises its predicted positive-class probability and moving it to a lower bucket lowers it, regardless of the transaction's true label, even though the buckets' training-data positive rates are nearly identical.

\begin{table}[t]
    \centering
    \caption{Training-set positive rate for the 10 dollar-buckets (ordered by increasing bucket). Buckets are encoded as Roman numerals in their true increasing order since the actual dollar ranges cannot be disclosed; bucket \textbf{Unknown} is a non-numeric category outside this ordering. Negative-class records were downsampled, so these rates are not representative of the true population rate.}
    \label{tab:dollar-bucket-stats-subset}
    \footnotesize
    \renewcommand{\arraystretch}{1.12}
    \resizebox{\linewidth}{!}{%
    \begin{tabular}{l*{10}{r}}
        \toprule
        \textbf{Dollar Bucket} & Unknown & I & II & III & IV & V & VI & VII & VIII & IX \\
        \midrule
        \textbf{Positive Rate} & 0.00\% & 31.65\% & 33.01\% & 29.70\% & 31.50\% & 31.07\% & 32.74\% & 36.83\% & 34.89\% & 32.44\% \\
        \bottomrule
    \end{tabular}%
    }
\end{table}

\section{Additional Analysis of Frontier Models}
\label{appdix:frontier_model}

\begin{figure}[t]
    \centering
    \includegraphics[width=\linewidth]{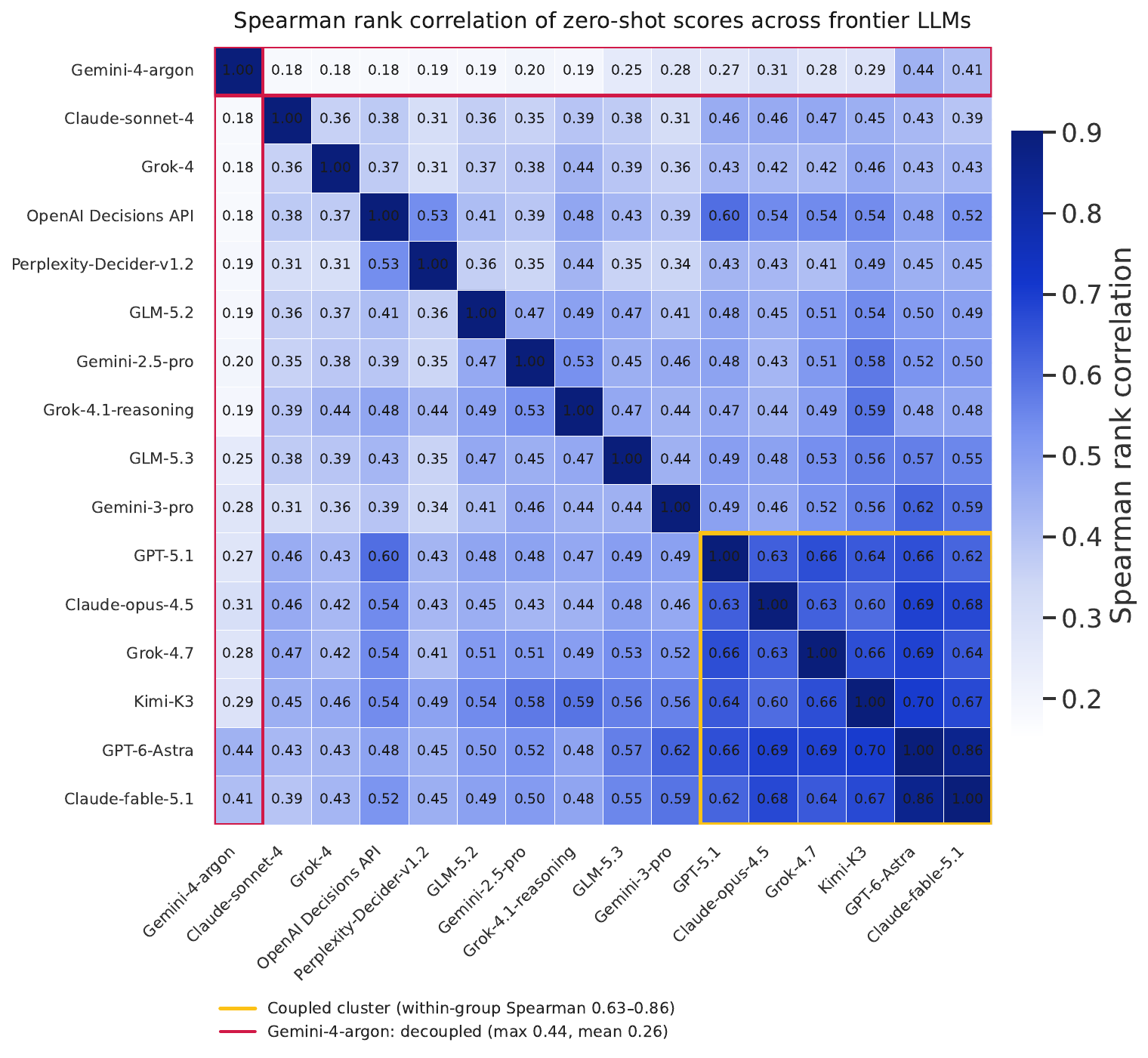}
    \caption{Spearman rank correlation of zero-shot scores across frontier LLMs on the 10K transaction classification sample. Models are ordered by hierarchical clustering on $1 - \rho$
   so related models appear as contiguous blocks. The gold box marks a tightly coupled cluster whose within-group correlations range from $0.63$ to $0.86$. The red outline highlights Gemini-4-argon, whose outputs are the most decoupled from the rest (mean correlation $0.26$, maximum $0.44$).}
    \label{fig:model_corr}
\end{figure}

We compute pairwise Spearman rank correlations between the output scores of frontier models evaluated in Sec.~\ref{sec:zeroshot_exp}.
According to Fig.~\ref{fig:model_corr}, the pairwise Spearman rank correlation scores span $0.18$ to $0.86$ and partition the models into two groups.
GPT-6-Astra, Claude-fable-5.1, Claude-opus-4.5, Kimi-K3, and Grok-4.7 form a tightly coupled cluster with within-cluster correlations between $0.63$ and $0.86$. Outside this cluster, outputs decouple sharply: Gemini-4-argon averages only $0.26$ correlation with the remaining listed models and reaches at most $0.44$ (with GPT-6-Astra). The concentration of high pairwise correlations among the top-ranked models suggests that they share some common properties that drive similar output score distributions.

\section{Open Source/Weight Model Source and License}

The source code, checkpoints, and license information of all open source/weight models we compared and used in the paper are listed in Table~\ref{tab:model-sources}.

\begin{table}[t]
    \centering
    \caption{Open source/weight models used in the paper and their code/checkpoint sources and licenses.}
    \label{tab:model-sources}
    \small
    \setlength{\tabcolsep}{4pt}
    \renewcommand{\arraystretch}{1.15}
    \begin{tabularx}{\linewidth}{
        >{\raggedright\arraybackslash\bfseries}p{0.18\linewidth}
        >{\raggedright\arraybackslash}X
        >{\raggedright\arraybackslash}p{0.39\linewidth}
    }
        \toprule
        Model & \textbf{Model/checkpoint source URL} & \textbf{License} \\
        \midrule
        XGBoost
            & \href{https://github.com/dmlc/xgboost}{dmlc/xgboost}
            & Apache-2.0.(\href{https://github.com/dmlc/xgboost/blob/master/LICENSE}{link}) \\
        LightGBM
            & \href{https://github.com/lightgbm-org/LightGBM}{lightgbm-org/LightGBM}
            & MIT (\href{https://github.com/lightgbm-org/LightGBM/blob/master/LICENSE}{link}) \\
        DCNv2
            & We implemented it based on the \href{https://dl.acm.org/doi/fullHtml/10.1145/3442381.3450078}{paper}.
            & CC-BY 4.0 (\href{https://dl.acm.org/doi/fullHtml/10.1145/3442381.3450078}{link}) \\
        SAINT
            & \href{https://github.com/jrzaurin/pytorch-widedeep}{pytorch-widedeep}
            & Apache-2.0 (\href{https://github.com/jrzaurin/pytorch-widedeep/blob/main/LICENSE_APACHE_v2_0}{link}) \\
        TabTransformer
            & \href{https://github.com/jrzaurin/pytorch-widedeep}{pytorch-widedeep}
            & Apache-2.0 (\href{https://github.com/jrzaurin/pytorch-widedeep/blob/main/LICENSE_APACHE_v2_0}{link}) \\
        TabPFN-3
            & \href{https://huggingface.co/Prior-Labs/tabpfn_3}{Prior-Labs/tabpfn\_3} --- tabpfn-v3-classifier-v3\_20260417\_binary.ckpt
            & TabPFN-3 License v1.0 (\href{https://huggingface.co/Prior-Labs/tabpfn_3/blob/main/LICENSE}{link}) \\
        TabFM
            & \href{https://huggingface.co/google/tabfm-1.1.0-pytorch}{google/tabfm-1.1.0-pytorch} --- classification/model.safetensors
            & TabFM Non-Commercial License v1.0 (\href{https://huggingface.co/google/tabfm-1.0.0-pytorch/blob/main/LICENSE}{link}) \\
        TabICLv2
            & \href{https://huggingface.co/jingang/TabICL}{jingang/TabICL} --- tabicl-classifier-v2-20260212.ckpt
            & BSD-3-Clause (\href{https://huggingface.co/jingang/TabICL}{github.com}) \\
        KumoTabular
            & \href{https://huggingface.co/nvidia/Kumo-Tabular}{nvidia/Kumo-Tabular} --- large/classifier.pt
            & OpenMDW-1.1 (\href{https://huggingface.co/nvidia/Kumo-Tabular/blob/main/LICENSE}{link}) \\
        Perplexity-Decider-v1.2
            & \href{https://huggingface.co/perplexity-ai/pplx-decider-v1.2}{perplexity-ai/pplx-decider-v1.2}
            & Apache-2.0(\href{https://huggingface.co/perplexity-ai/pplx-decider-v1.2}{link}) \\
        BERT
            & \href{https://huggingface.co/google-bert/bert-base-uncased}{google-bert/bert-base-uncased}
            &Apache-2.0 (\href{https://huggingface.co/google-bert/bert-base-uncased}{link}) \\
        NV-Embed-v2
            & \href{https://huggingface.co/nvidia/NV-Embed-v2}{nvidia/NV-Embed-v2}
            & CC BY-NC 4.0 (\href{https://huggingface.co/nvidia/NV-Embed-v2}{link}) \\
        Llama 3.2
            & \href{https://huggingface.co/meta-llama/Llama-3.2-1B}{meta-llama/Llama-3.2-1B}\newline
              \href{https://huggingface.co/meta-llama/Llama-3.2-3B}{meta-llama/Llama-3.2-3B}
            & LLAMA 3.2 COMMUNITY LICENSE (\href{https://huggingface.co/meta-llama/Llama-3.2-1B/blob/main/LICENSE.txt}{link}) \\
        MobileLLM
            & \href{https://huggingface.co/facebook/MobileLLM-125M}{facebook/MobileLLM-125M}
            & FAIR Noncommercial Research License (\href{https://huggingface.co/facebook/MobileLLM-125M/blob/main/LICENSE}{link}) \\
        \bottomrule
    \end{tabularx}
\end{table}

\end{document}